\documentclass[10pt,twocolumn,letterpaper]{article}

\usepackage{wacv}
\usepackage{times}
\usepackage{epsfig}
\usepackage{graphicx}
\usepackage{amsmath}
\usepackage{amssymb}
\usepackage{booktabs}
\usepackage[T1]{fontenc}

\usepackage{lipsum}
\usepackage{bm}
\usepackage{relsize} 
\usepackage{booktabs, multirow, array, makecell}
\usepackage{soul}
\usepackage{subcaption}
\usepackage{wrapfig}

\usepackage{floatrow}
\newfloatcommand{capbtabbox}{table}[][\FBwidth]

\usepackage{mathrsfs}

\usepackage{pifont}
\newcommand{\cmark}{\ding{51}}

\usepackage{comment}
\usepackage{color}

\usepackage{authblk}

\def\wacvPaperID{702} % Enter the WACV Paper ID here

\wacvalgorithmstrack   % Uncomment this line if you are submitting to the Algorithms Track.
\wacvfinalcopy % *** Uncomment this line for the final submission

\ifwacvfinal
\usepackage[breaklinks=true,bookmarks=false]{hyperref}
\else
\usepackage[pagebackref=true,breaklinks=true,colorlinks,bookmarks=false]{hyperref}
\fi

\begin{document}

%%%%%%%%% TITLE

\title{Uncertainty-aware Label Distribution Learning for \\ Facial Expression Recognition} 

% \author{First Author\\
% Institution1\\
% Institution1 address\\
% {\tt\small firstauthor@i1.org}
% % For a paper whose authors are all at the same institution,
% % omit the following lines up until the closing ``}''.
% % Additional authors and addresses can be added with ``\and'',
% % just like the second author.
% % To save space, use either the email address or home page, not both
% \and
% Second Author\\
% Institution2\\
% First line of institution2 address\\
% {\tt\small secondauthor@i2.org}
% }

\author[1, 2, 3]{Nhat Le$^*$}
\author[1, 2, 5]{Khanh Nguyen$^*$}
\author[3]{Quang Tran}
\author[3]{Erman Tjiputra}
\author[1, 2]{Bac Le}
\author[4]{Anh Nguyen}

\affil[1]{\small Faculty of Information Technology, University of Science, Ho Chi Minh City, Vietnam}
\affil[2]{\small Vietnam National University, Ho Chi Minh City, Vietnam}
\affil[3]{\small AIOZ, Singapore}
\affil[4]{\small Department of Computer Science, University of Liverpool, Liverpool, UK}
\affil[5]{\small FPT Software AI Center, Vietnam}

\maketitle
\def\thefootnote{*}\footnotetext{Equal contribution}

\thispagestyle{empty}

\begin{abstract}
Despite significant progress over the past few years, ambiguity is still a key challenge in Facial Expression Recognition (FER). It can lead to noisy and inconsistent annotation, which hinders the performance of deep learning models in real-world scenarios. In this paper, we propose a new uncertainty-aware label distribution learning method to improve the robustness of deep models against uncertainty and ambiguity. We leverage neighborhood information in the valence-arousal space to adaptively construct emotion distributions for training samples. We also consider the uncertainty of provided labels when incorporating them into the label distributions. Our method can be easily integrated into a deep network to obtain more training supervision and improve recognition accuracy. Intensive experiments on several datasets under various noisy and ambiguous settings show that our method achieves competitive results and outperforms recent state-of-the-art approaches. Our code and models are available at \url{https://github.com/minhnhatvt/label-distribution-learning-fer-tf}.

%Facial expression recognition is a crucial task to understand human emotions. Most recent works focus on improving recognition accuracy, while real-world problems such as noisy labels or ambiguity are not fully investigated. In this paper, we propose a new label distribution learning approach to improve the robustness of facial expression recognition methods when dealing with noisy or ambiguous data. Our work utilizes the valence-arousal to represent human emotions in continuous space, then establishes the relationship between image instances to construct emotion distributions. The constructed distribution can be integrated into a deep network to provide more supervision training signals and improve the recognition accuracy. Intensive experiments on large-scale datasets under various settings show that our method outperforms recent state-of-the-art approaches, while being robust against noisy and inconsistent labels. 
\end{abstract}

\section{Introduction}
\label{sec:intro}
Facial expression recognition (FER) plays an important role in understanding people's feelings and interactions between humans. Recently, automatic emotion recognition has gained a lot of attention from the research community~\cite{academia_industry} due to its applications in healthcare~\cite{application_health_care}, surveillance~\cite{application_surveillance}, or human-robot interaction~\cite{application_hci}. Most recent FER methods utilize deep learning~\cite{cnn} and achieve better results than handcrafted features approaches \cite{hog,lbp}. The success of deep networks can be attributed to large-scale FER datasets such as AffectNet \cite{affectnet}, EmotioNet \cite{emotionet}, and RAF-DB \cite{raf_db}. Some datasets describe emotion in terms of Action Units (AUs) following the Facial Action Coding System \cite{facs} or quantify affection over continuous scales, such as valence and arousal \cite{dim_model}, while most of them classify facial expressions into basic universal emotions \cite{6emotions,contempt} and the neutral state.
% including \texttt{anger}, \texttt{disgust}, \texttt{fear}, \texttt{happiness}, \texttt{sadness}, \texttt{surprise},  \texttt{neutral} \cite{6emotions}, and \texttt{contempt} \cite{contempt}. 

\begin{figure*}[t]
    \centering
    \includegraphics[width=0.9\textwidth]{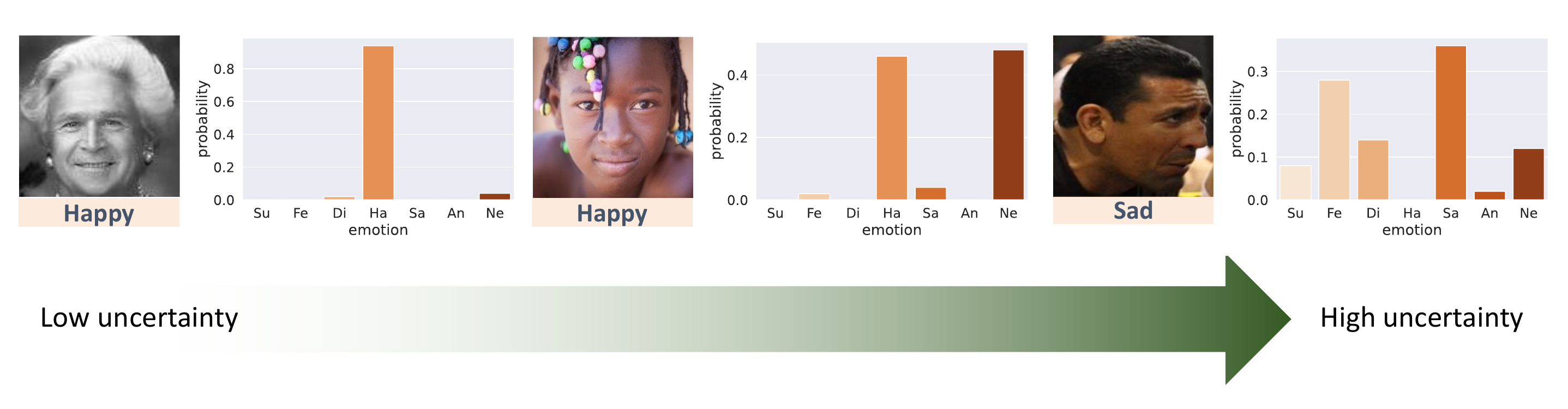}
    \caption{
    % \vspace{-2ex}
    User study results by 50 volunteers on three random images from RAF-DB dataset. The expression on the right image are more ambiguous, which leads to high uncertainty in the emotion label. Labels at the bottom denote the provided annotation from the dataset. (Su=\texttt{Surprise}, Fe=\texttt{Fear}, Di=\texttt{Disgust}, Ha=\texttt{Happy}, Sa=\texttt{Sad}, An=\texttt{Angry}, Ne=\texttt{Neutral})
    %Also, a single-label provided may not  cover sufficient emotional information to train a neural network model efficiently.
    }
    \vspace{-3ex}
    \label{fig:motivation}
\end{figure*}

Unfortunately, large-scale FER datasets often suffer from the problem of label uncertainty and annotation ambiguity \cite{ipa2lt,ldl_alsg,dmue}. People with different backgrounds might perceive and interpret facial expressions differently, which can lead to inconsistent and uncertain labels \cite{ipa2lt,dmue}. In addition, real-life facial expressions usually manifest a mixture of feelings \cite{edl,ldl_alsg} rather than a single exaggerated emotion often found in the lab-controlled setting.
For example, Figure~\ref{fig:motivation} shows that people may have different opinions about the expressed emotion, particularly in ambiguous images. Consequently, a distribution over emotion categories is better than a single label because it takes all sentiment classes into account and can cover various interpretations, thus mitigating the effect of ambiguity \cite{deepldl}. However, most current large-scale FER datasets only provide a single label for each sample instead of a label distribution, which means we do not have a comprehensive description for each facial expression. This can lead to insufficient supervision during training and pose a big challenge for many FER systems.

To overcome annotation ambiguity in FER, this paper proposes a new uncertainty-aware label distribution learning method that constructs emotion distributions for training samples. Specifically, for each instance, we leverage valence-arousal information to identify a set of neighbors and calculate their corresponding contributions using our adaptive similarity mechanism. We then aggregate neighborhood information with the provided single label, adjusted by its learnable uncertainty factor, to generate the target label distribution. Finally, we use the constructed distribution as supervision signals to optimize the model via label distribution learning. We also introduce a discriminative loss that reduces intra-class variations and encourages inter-class differences to improve the model's robustness against ambiguous features. Note that the distribution construction only occurs during training while the inference process remains intact. 
% The extensive experiments show that our method outperforms recent state-of-the-arts and achieves competitive results, both quantitatively and qualitatively. 
In summary, our contributions are as follows:
\begin{enumerate}
    \item We propose a new method, namely \textbf{L}abel \textbf{D}istribution \textbf{L}earning with \textbf{V}alence-\textbf{A}rousal (LDLVA), for FER with ambiguous annotation by exploiting neighborhood information in the valence-arousal space.
    % Our method can be easily integrated into existing networks to improve their robustness against ambiguity without introducing extra computation costs to inference.
    % without changing the backbone.
    % The target label distribution is constructed with the assistance of valence-arousal space.
    \item Our uncertainty-aware label distribution construction provide more accurate and richer supervision for training deep FER networks, allowing them to learn from ambiguous data effectively in an end-to-end manner. 
    % \item To combat against ambiguous features, a modification of center loss is designed to enhance the learned representations by reducing intra-class variations and increasing inter-class differences.
    \item We perform extensive experiments under various synthetic and real-world ambiguity settings and achieve state-of-the-art results on RAF-DB, AffectNet, and SFEW datasets.    
\end{enumerate}

% \begin{figure*}[t]
%     \centering
%     \includegraphics[width=0.95\textwidth]{figs/motivation.pdf}
%     \caption{
%     %\vspace{-2ex}
%     User study results by 50 volunteers on three random images from RAF-DB dataset. The expression on the right images are more ambiguous, which leads to high uncertainty in emotion labels. Labels at the bottom denote the provided annotation from the dataset. (Su=\texttt{Surprise}, Fe=\texttt{Fear}, Di=\texttt{Disgust}, Ha=\texttt{Happy}, Sa=\texttt{Sad}, An=\texttt{Angry}, Ne=\texttt{Neutral})
%     %Also, a single-label provided may not  cover sufficient emotional information to train a neural network model efficiently.
%     }
%     \vspace{-3ex}
%     \label{fig:motivation}
% \end{figure*}
% \vspace{-2ex}
\section{Related Work}
\label{sec:related_works}
Most recent methods~\cite{deexpression_res,cake,fsn,dacl_net,ldl_alsg,scn,ipa2lt,caer,zhang2022eac,psr,glamor_net} categorize facial expression into discrete classes corresponding to basic universal emotions~\cite{6emotions,contempt}, which is easy to interpret and intuitive to humans. Other approaches~\cite{joint_AU,unmask_FAC} attempted to represent human emotion using Action Units (AUs)~\cite{facs} or continuous scales such as valence and arousal~\cite{dim_model}. In this work, we leverage the auxiliary information of continuous scales to mitigate the effect of uncertainty and ambiguity in existing FER datasets when predicting the discrete emotion of a given facial expression.

% Although several progress have been made to improve the performance on FER based on deep learning, the problem of making deep networks robust to uncertain and ambiguous label is not fully investigated.
One challenging problem of FER is that ambiguous facial expressions can make it difficult to correctly identify the expressed emotions, which might lead to noisy and uncertain annotations~\cite{ldl_alsg,dmue}. Empirical studies also show that neural networks are sensitive to noise and can easily overfit noisy data \cite{dnn_overfit,ldl_joint_optim,ldl_pencil}. To overcome this challenge, previous approaches model the noise by a transition matrix~\cite{sukhbaatar,nal,patrini,ipa2lt}.
% The authors in~\cite{nal} adopt a transition probability layer on top of the network prediction and train the whole model end-to-end.
In~\cite{air}, precise image features are extracted from a pre-trained model to regulate the learning process with noisy labels. The authors in~\cite{robust_loss1,robust_loss2} use noise-tolerant loss functions to increase noise robustness. Other methods~\cite{meta_cleaner,scn} measure the uncertainty of each sample and utilize a sample-weighting strategy to help the network tolerate noisy samples. Recently, Zhang \etal~\cite{rul} propose to quantify the uncertainties from the relative difficulty of samples by feature mixup.
However, these methods only focus on improving the accuracy on mislabelled data and do not handle the ambiguous nature of facial expressions.

An alternative approach to address label noise and ambiguity is label distribution learning (LDL)~\cite{ldl}. In other domains, previous works~\cite{head_pose,facial_age,deepldl} leverage prior knowledge to transfer logical labels into discretized bivariate Gaussian label distribution. The authors in~\cite{ldl_joint_optim,ldl_pencil} utilize network's predictions as label
distributions to correct noise, which can be unstable and hard to optimize. Instead, our method not only adaptively utilizes the model's predictions but also exploits domain knowledge of the valence-arousal space to construct target distributions. 
In FER literature, Zhou \etal~\cite{edl} introduces a framework to map a facial expression to multiple emotions with corresponding intensities. Jia \etal~\cite{local_ldl} proposes to learn emotion distributions by exploiting label correlations at a local level. Zhao \etal~\cite{lightweight_ldl} uses a pre-trained label distribution generator to produce emotion distribution. Other works create the label distribution by computing the membership degrees to the labels~\cite{soft_membership,fuzzy_membership,soft_facial_landmark,labeling_importance,manifold_learning}. Recently, Chen \etal~\cite{ldl_alsg} leverages the topology in facial landmarks and action units spaces to acquire more information for label distribution learning. She \etal~\cite{dmue} proposed to leverage multiple branches to obtain the latent distribution. However, these methods either rely heavily on good features with local linearity to work properly ~\cite{soft_membership,do2022fine,fuzzy_membership,soft_facial_landmark,labeling_importance,nguyen2021graph,manifold_learning} or only use the mined label distributions to regularize the model's training process instead of directly learning from them~\cite{ldl_alsg,dmue}.
%These methods are based on strong assumptions such as the local linearity in the feature space and often require good features to work properly.%, which might be inappropriate for features that are extracted from deep networks.
% Recently, Chen \etal~\cite{ldl_alsg} leverages the topological information in the facial landmarks and action units space to obtain more information for the label distribution learning process. She \etal~\cite{dmue} proposed to leverage multiple branches to obtain the latent distribution. However, \cite{ldl_alsg,dmue} only used the mined label distributions to guide the target branch towards the mined information. %Their main training objective is the standard cross-entropy between the network prediction and the provided single label. 

Unlike previous works, our method constructs emotion distributions for training instances and directly uses them as supervision information, thus reducing the effects of annotation ambiguity. 
We do not need to be provided with label distributions to train the network since they can be accurately estimated using our adaptive similarity mechanism and learnable uncertainty factors.
We experimentally show that our approach is more effective as the network is trained end-to-end with label distributions, which brings more meaningful information to the training process. 

\section{Methodology}
\begin{figure*}[h]
    \vspace{-2ex}
    \centering
    \includegraphics[width=0.9\textwidth]{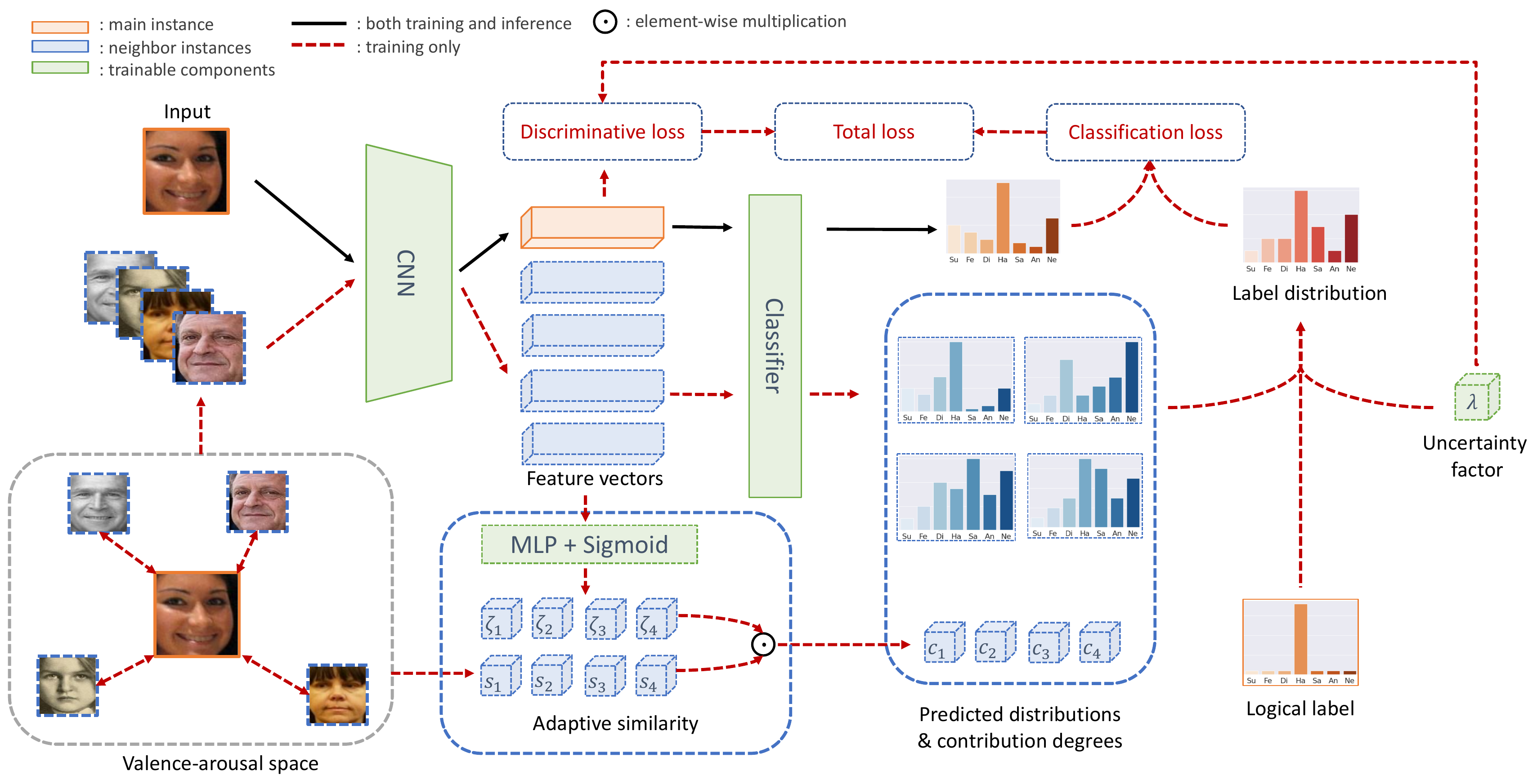}
    \caption{An overview of our Label Distribution Learning with Valence-Arousal (LDLVA) for facial expression recognition under ambiguity. Dotted lines denote components used in training only while solid lines denote components used in both training and testing.}
    \vspace{-2ex}
    \label{fig:overall_framework}
\end{figure*}
We first introduce a list of notations that will be used throughout this paper. Let $\bm{x} \in \mathcal{X}$ be the instance variable in the input space $\mathcal{X}$ and $\bm{x}^{i}$ be the particular $i$-th instance.
The label set is denoted as $\mathcal{Y} = \{y_1, y_2,..., y_m\}$ where $m$ is the number of classes and $y_j$ is the label value of the $j$-th class. 
The logical label vector of $\bm{x}^{i}$ is indicated by $\bm{l}^{i}$ = $(l^{i}_{y_1}, l^{i}_{y_2}, ..., l^{i}_{y_m})$ with $\l^{i}_{y_j} \in \{0, 1\}$ and $\| \bm{l} \| _1 = 1$. 
We define the label distribution of $\bm{x}^{i}$ as $\bm{d}^{i}$ = $(d^{i}_{y_1}, d^{i}_{y_2}, ..., d^{i}_{y_m})$ with $\| \bm{d} \| _1 = 1$ and $d^{i}_{y_j} \in [0, 1]$ representing the relative degree that $\bm{x}^{i}$ belongs to the class $y_j$. 
% Correspondingly, a $n$-sample \textit{training} set with logical and distribution labels are denoted as $D_l$ = $\{ (\bm{x}^{i}, \bm{l}^{i}) \vert 1 \le i \le n\}$ and $D_l$ = $\{ (\bm{x}^{i}, \bm{l}^{i}) \vert 1 \le i \le n\}$. 
A neural network with parameters $\theta$ followed by a softmax layer is denoted as $f(\bm{x}; \theta)$.
The corresponding feature vector of $\bm{x}^i$ extracted by a CNN backbone model is indicated by $\bm{v}^i \in \mathbb{R}^V$.

\subsection{Overview}
Most existing FER datasets assign only a single class or equivalently, a logical label $\bm{l}^{i}$ for each training sample $\bm{x}^{i}$. In particular, the given training dataset is a collection of $n$ samples with logical labels $D_l$ = $\{ (\bm{x}^{i}, \bm{l}^{i}) \vert 1 \le i \le n\}$. However, as depicted in Figure \ref{fig:motivation}, a label distribution $\bm{d}^i$ is a more comprehensive and suitable annotation for the image than a single label. 
Inspired by the recent success of label distribution learning (LDL) in addressing label ambiguity \cite{deepldl}, we aim to construct an emotion distribution $\bm{d}^i$ for each training sample $\bm{x}^i$, thus transform the \textit{training} set $D_l$ into $D_d$ = $\{ (\bm{x}^{i}, \bm{d}^{i}) \vert 1 \le i \le n\}$, which can provide richer supervision information and help mitigate the ambiguity issue. Consequently, our goal is to optimize the parameters $\theta$ of the neural network $f(\bm{x}; \theta)$ such that it can learn an appropriate mapping function for the instance $\bm{x}^i$ from the input space to the target label distribution $\bm{d}^i$. Mathematically, we use cross-entropy to measure the discrepancy between the model's prediction and the constructed target distribution \cite{deepldl}. Hence, the solution can be obtained by minimizing the following classification loss:
\begin{equation}
    \label{eq:classification_loss}
    \mathcal{L}_{cls} = \sum_{i=1}^n \text{CE}\left(\bm{d}^i, f(\bm{x}^i; \theta)\right)
    = -\sum_{i=1}^n \sum_{j=1}^m \bm{d}_j^{i} \log f_j(\bm{x}^{i};\theta).
\end{equation}

An overview of our method is presented in Figure \ref{fig:overall_framework}. To construct the \textit{label distribution} for each {training} instance $\bm{x}^i$, we leverage its neighborhood information in the valence-arousal space. Particularly, we identify $K$ neighbor instances for each training sample $\bm{x}^i$ and utilize our \textit{adaptive similarity mechanism} to determine their contribution degrees to the target distribution $\bm{d}^i$. Then, we combine the neighbors' predictions and their corresponding contribution degrees with the provided label $\bm{l}^i$ and $\bm{l}^i$'s uncertainty factor to obtain the label distribution $\bm{d}^i$. The constructed distribution $\bm{d}^i$ will be used as supervision information to train the model via label distribution learning.
% for the standard label distribution learning to optimize the model's parameters. 
% The overall process consists of three main steps: 1) utilize our \textit{adaptive similarity mechanism} to determine the contribution of each neighbor to the target distribution $\bm{d}^i$; 2) incorporate \textit{uncertainty factor} $\lambda$ with logical label $\bm{l}^i$ to mitigate the adverse effect of ambiguous pseudo annotation, together with neighborhood information to construct the label distribution $\bm{d}^i$; 3) optimize trainable parameters through \textit{classification} and \textit{discriminative} loss.
It is worth noting that these steps occur only during training, thus no extra costs are introduced at inference time. 
% \textbf{3 steps explanation is a bit confusing} %More details about the process will be discussed in the following sections. 

\subsection{Adaptive Similarity Measuring}
As in previous works \cite{smoothness_assumption,le_ldl,ldl_alsg}, we assume that facial images should have similar emotions to their neighbors in an auxiliary or supporting space. Therefore, the label distribution of an instance can be constructed using the information of its neighbors. Since our goal is to reconstruct the target label distribution with high fidelity, the chosen supporting space should highly correlate with the emotion space to transfer as much information as possible.
Although information such as facial landmarks and action units can be utilized as the supporting space, we find that valence-arousal values are more closely associated with discrete emotions and thus particularly suitable to be the auxiliary space. In practice, the valence-arousal has been widely used to represent the human emotional spectrum, with valence describing how positive or negative an expression is and arousal indicating the intensity or activation degree of the expression \cite{valence_arousal}.

Similar to the smoothness assumption \cite{smoothness_assumption}, we assume that the label distribution of the main instance $\bm{x}^i$ can be computed as a linear combination of its neighbors' distributions. 
To determine the contribution of each neighbor, we propose an adaptive similarity mechanism that not only leverages the relationships between $\bm{x}^i$ and its neighbors in the auxiliary space but also utilizes their feature vectors extracted from the backbone. 
% We first use the $K$-Nearest Neighbor algorithm to identify $K$ closest points for each training sample $\bm{x}^i$, denoted as $N(i)$, based on the distance between training instances in the valence-arousal space.
In particular, we first use the $K$-Nearest Neighbor algorithm to identify $K$ closest points for each training sample $\bm{x}^i$, denoted as $N(i)$, based on the distance between training instances in the valence-arousal space. We then compute a \textit{local similarity score} between $\bm{x}^i$ and each of its $K$ neighbors using the following formula:
% in \cite{le_ldl}:
\begin{equation}
    \label{eq:similarity_score}
    s^i_k = \exp\left(-\frac{\| \bm{a}^i - \bm{a}^k \|^2_2}{\delta^2}\right), \quad \forall \bm{x}^k \in N(i),
\end{equation} 
where $\bm{a}$ is the corresponding auxiliary valence-arousal vector of $\bm{x}$, and $\delta$ is a hyperparameter controlling similarity measurement. Intuitively, the higher $s^i_k$ is, the more $\bm{x}^k$ contributes to the label distribution of $\bm{x}^i$.

% We find that some neighbors of an instance may not contribute well to the estimation of the main instance’s label distribution. Assigning uniform contributions, i.e., every neighbor has equal weight, may lead to inaccurate distribution construction. Therefore, our adaptive similarity mechanism is designed to adaptively measure the importance of each neighbor instance.
However, since valence-arousal values are not always available in practice, we leverage an existing method {\cite{va_prediction}} to generate pseudo-valence-arousal. Consequently, these values can be inaccurate and lead to incorrect calculation of $s^i_k$. Therefore, we proposed to correct these potential errors with our adaptive similarity mechanism.
Specifically, we calculate a \textit{calibration score} for each $(\bm{x}^i, \bm{x}^k)$ pair using the feature vectors $(\bm{v}^i, \bm{v}^{k})$ extracted by the CNN backbone of $\bm{x}^i$ and its neighbor instance $\bm{x}^k \in N(i)$ as follows:
\begin{equation}
    \label{eq:mlp}
    \zeta^i_k = \text{Sigmoid}\left(g([\bm{v}^i, \bm{v}^{k}];\phi)\right),
\end{equation}
where $[\boldsymbol{\cdot}, \boldsymbol{\cdot}]$ is the concatenation operator, $g$ is a three-layer perceptron (MLP) with parameter $\phi$. The dimensionality of each layer is 512, 256, and 1, respectively. We also apply layer normalization and ReLU non-linearity in the first two layers.

The final \textit{contribution degrees} of neighbor instances are calculated as the product of the local similarity and the calibration score:
\begin{equation}
    \label{eq:contribution_factor}
    c^i_k = 
    \begin{cases}
    \zeta^i_k  s^i_k, &\text{for } \bm{x}^k \in N(i), \\
    0,  &\text{otherwise}.
    \end{cases}
\end{equation}

\subsection{Uncertainty-aware Label Distribution Construction}
After obtaining the contribution degree of each neighbor $\bm{x}^k \in N(i)$, we can now generate the target label distribution $\bm{d}^i$ for the main instance $\bm{x}^i$. The target label distribution is calculated using the logical label $\bm{l}^i$ and the aggregated distribution $\Tilde{\bm{d}}^i$ defined as follows:
\begin{align}
    \label{eq:aggregated_distribution}
    \Tilde{\bm{d}^i} &= \frac{\sum_k c^i_k  f(\bm{x}^{k};\theta)}{\sum_k c^i_k}, \\
    \label{eq:target_distribution}
    \bm{d}^i &= (1-\lambda^i) \bm{l}^i + \lambda^i \Tilde{\bm{d}^i},
\end{align}
where $\lambda^i \in [0,1]$ is the \textit{uncertainty factor} for the logical label. It controls the balance between the provided label $\bm{l}^i$ and the aggregated distribution $\Tilde{\bm{d}^i}$ from the local neighborhood. Intuitively, a high value of $\lambda^i$ indicates that the logical label is highly uncertain, which can be caused by ambiguous expression or low-quality input images as illustrated in Figure \ref{fig:uncertainty}, thus we should put more weight towards neighborhood information $\Tilde{\bm{d}^i}$. Conversely, when $\lambda^i$ is small, the label distribution $\bm{d}^i$ should be close to $\bm{l}^i$ since we are certain about the provided manual label. In our implementation, $\lambda^i$ is a trainable parameter for each instance and will be optimized jointly with the model's parameters using gradient descent. 

%we assign each training instance a separate trainable value $\hat{\lambda}^i$ which 
% The uncertainty factor used in Equation \ref{eq:target_distribution} can be obtained by putting $\hat{\lambda}^i$ through a sigmoid function.
% $\lambda^i = \text{Sigmod}(\hat{\lambda}^i)$.
% Next, we provide the explanation about our training strategy in which the uncertainty factors are iteratively updated.

% ===================
% put it in sup?

Mathematically speaking, consider Equation \ref{eq:classification_loss} and \ref{eq:target_distribution}, the derivative of $\mathcal{L}_{cls}$ with respect to $\lambda^i$ can be computed as:
\begin{align}
    \frac{\partial \mathcal{L}_{cls}}{\partial\lambda^i} 
    &= \frac{\partial \text{CE}\left( \bm{d}^i, f(\bm{x}^i; \theta) \right)}{\partial\lambda^i} \\
    % &= \frac{\partial \left[ -\sum_j \left((1-\lambda^i)\bm{l}^i_j + \lambda^i\Tilde{\bm{d}^i_j}\right)\log f_j(\bm{x}^i; \theta) \right]}{\partial\lambda^i} \\ 
    % &= -\sum_j (\Tilde{\bm{d}^i_j} - \bm{l}^i_j)\log f_j(\bm{x}^i; \theta) \\
    &= -\sum_j\Tilde{\bm{d}^i_j}\log f_j(\bm{x}^i; \theta)  + \sum_j\bm{l}^i_j\log f_j(\bm{x}^i; \theta) \\
    &= \text{CE}(\Tilde{\bm{d}^i}, f(\bm{x}^i; \theta)) - \text{CE}(\bm{l}^i,  f(\bm{x}^i; \theta)).
\end{align}
If $\text{CE}(\bm{l}^i, f(\bm{x}^i; \theta))$ is smaller than $\text{CE}(\Tilde{\bm{d}^i},f(\bm{x}^i; \theta))$, the derivative of $\mathcal{L}_{cls}$ with respect to $\lambda^i$ is positive, which leads to a negative update for $\lambda^i$ following gradient descent optimization scheme. This is desirable because in this case, the network output is in more agreement with the logical label than the aggregated neighborhood distribution. In other words, it is more confident about the provided label and thus, we should decrease the value of the uncertainty factor $\lambda^i$. The same reasoning can be applied in the opposite situation.

\subsection{Loss Function}
% To improve the ability of the model to recognize facial expression more precisely, recent literatures have shown the benefits of learning discriminative features \cite{island_loss,affwild2,dda_loss,dacl_net}. 
Recent literatures have shown the benefits of learning discriminative features in FER \cite{island_loss,affwild2,dda_loss,dacl_net}. Inspired by this, we believe it is beneficial to encourage the network to learn good facial descriptions because it can help improve the model's ability to discriminate between ambiguous emotions.
We find that the center loss \cite{center_loss} is suitable for our purpose because of its simplicity and efficacy in reducing the intra-class variations of the learned representations. 
Nevertheless, in the traditional formulation of the center loss \cite{center_loss}, the features of a sample are ``blindly" pulled towards its corresponding class center given its label. This means when the provided label is incorrect, it can cause the network to learn imprecise features. We propose to overcome this problem by incorporating the label uncertainty factor $\lambda^i$ to adaptively penalize the distance between the sample and its corresponding center. For instances with high uncertainty, the network can effectively tolerate their features in the optimization process. 
Furthermore, we also add pairwise distances between class centers to encourage large margins between different classes, thus enhancing the discriminative power. Our discriminative loss is calculated as follows:
% In addition to the classification loss introduced in Equation  \ref{eq:classification_loss}, we also employ a discriminative loss calculated as follows:
% \begin{align}
% \label{eq:discriminative_loss}
%     \mathcal{L}_D &= \frac{1}{2}\sum_{i=1}^n (1-\lambda^i)\Vert \bm{v}^i - \bm{\mu}_{y^i} \Vert_2^2 \notag\\
%     &+ \sum_{j=1}^m \sum_{\substack{k=1 \\ k \neq j}}^m \exp \left(-\frac{\Vert\bm{\mu}_{y_j}-\bm{\mu}_{y_k}\Vert_2^2}{\sqrt{V}}\right),
% \end{align}
% where $y^i$ is the class label of the $i$-th sample, $\bm{\mu}_{y^i} \in \mathbb{R}^V$ is the class center for class $y^i$,  $\bm{\mu}_{y_j}$ and $\bm{\mu}_{y_k}$ are class center for class ${y_j}$ and ${y_k}$, respectively. 
\begin{align}
\label{eq:discriminative_loss}
    \mathcal{L}_D &= \frac{1}{2}\sum_{i=1}^n (1-\lambda^i)\Vert \bm{v}^i - \bm{\mu}_{y^i} \Vert_2^2 \notag\\
    &+ \sum_{j=1}^m \sum_{\substack{k=1 \\ k \neq j}}^m \exp \left(-\frac{\Vert\bm{\mu}_{j}-\bm{\mu}_{k}\Vert_2^2}{\sqrt{V}}\right),
\end{align}
where $y^i$ is the class index of the $i$-th sample while $\bm{\mu}_{j}$, $\bm{\mu}_{k}$, and $\bm{\mu}_{y^i}$ $\in \mathbb{R}^V$ are the center vectors of the ${j}$-th, ${k}$-th, and $y^i$-th classes, respectively.
During the training phase, all center vectors are zero-initialized and optimized using Equation \ref{eq:discriminative_loss}. %We do not back propagate gradients through $\lambda^i$ in this equation as it can make the uncertainty value drop to zeros quickly.
Intuitively, the first term of $\mathcal{L}_D$ encourages the feature vectors of one class to be close to their corresponding center \cite{center_loss} while the second term improves the inter-class discrimination by pushing the cluster centers far away from each other.

Combining Equation \ref{eq:classification_loss} and Equation \ref{eq:discriminative_loss}, we obtain the total loss for training:
\begin{equation}
    \label{eq:total_loss}
    \mathcal{L} = \mathcal{L}_{cls} + \gamma\mathcal{L}_D,
\end{equation}
where $\gamma$ is the hyperparameter balancing between the two losses.

\section{Experiments}
\label{sec:experiments}
In this section, we first validate the effectiveness of our approach on synthetic ambiguity caused by noisy label data. Next, we evaluate the performance of our LDLVA in handling inconsistent labels caused by ambiguous facial expressions. We then compare LDLVA with state-of-the-art methods to demonstrate the robustness of our approach towards annotation ambiguity that inherently exists in real-world data. Finally, we conduct ablation studies and present qualitative results to investigate the effectiveness of each component as well as the advantages of our method.

\subsection{Datasets}
We perform experiments on three popular in-the-wild FER datasets: AffectNet~\cite{affectnet}, RAF-DB~\cite{raf_db} and SFEW~\cite{sfew}. They are created by collecting data from the Internet and reflect real-life scenarios.
AffectNet \cite{affectnet} has more than 400,000 facial images manually annotated with discrete emotions and valence-arousal. Following previous work \cite{ldl_alsg,ipa2lt,dacl_net}, we select approximately 280,000 and 3,500 images for training and testing, all of which belong to six basic emotions (\texttt{surprise}, \texttt{fear}, \texttt{disgust}, \texttt{happy}, \texttt{sad}, and \texttt{angry}) and \texttt{neutral} expression. RAF-DB \cite{raf_db} is split into training and test sets with more than 12,000 and 3,000 images, respectively. SFEW \cite{sfew} has 879 training images and 406 testing images, all of which are extracted from movie videos.
% Images from RAF-DB and SFEW datasets are only annotated with categorical emotion without valence-arousal information. Therefore, we employ an existing model \cite{va_prediction} 
% trained on AffectNet to generate pseudo valence-arousal for RAF-DB and SFEW images.
% Since the AffectNet dataset has an imbalanced training set but a balanced test set, we use oversampling strategy to train our network.

% We follow the same split in other state-of-the-art methods \cite{ldl_alsg,ipa2lt,dacl_net} for training and testing. %In all datasets, the emotions are classified into $7$ classes: \texttt{angry}, \texttt{disgust}, \texttt{fear}, \texttt{happy}, \texttt{sad}, \texttt{surprise}, and \texttt{neutral}. 
% Since the valence-arousal values are not provided in RAF-DB and SFEW datasets, we employ the valance-arousal detection method \cite{va_prediction} trained on AffectNet to generate the pseudo valence-arousal for RAF-DB and SFEW datasets. 
 
% To verify the robustness of our method against uncertainty, we also conduct experiments by adding noise discussed in {\cite{patrini}} and {\cite{ldl_joint_optim}} to RAF-DB, AffectNet and SFEW datasets. The noise is created by randomly flipping the correct label $y$ to one of the other classes $\Tilde{y}$.
\subsection{Experimental Settings}
% By default, we use ResNet-18 \cite{resnet} pretrained on MS-Celeb-1M as the backbone and denote our corresponding network as LDLVA. Only in the experiment on original datasets (subsection \ref{sec:exp_original}), we use backbone ResNet-50 \cite{resnet} for a fair comparison with previous state-of-the-art methods that have complex feature extractors. 
By default, we use the pretrained ResNet-50 \cite{resnet} as the CNN backbone.
% for a fair comparison with previous works \cite{rul,scn}. 
% The output of the last pooling layer of the backbone is employed as the facial feature vector.
We align the input image and perform on-the-fly augmentation during training by randomly flipping the image horizontally and taking a random crop of size 224×224 after padding 16 pixels on each side. At test time, we use the central crop of the image as input for the model. During training, for each instance, we consider 8 nearest neighbors and initialize its uncertainty factor $\lambda^i$ to zero. To optimize the discriminative loss (Equation \ref{eq:discriminative_loss}), we follow the same settings as in \cite{center_loss}.
We train the network using Adam optimizer \cite{adam} with batch size 32 for 30 epochs with an initial learning rate of $0.001$. The parameters $\delta$ in Equation \ref{eq:similarity_score} and $\gamma$ in Equation \ref{eq:total_loss} are set to 0.5 and 0.1 based on validation results. Similar to previous works \cite{ldl_alsg,rul,scn}, we use the overall accuracy as the metric to evaluate the models.

\subsection{Experiments with Noisy Labels}

\begin{table}[h]
    \centering
    \setlength\aboverulesep{0pt}\setlength\belowrulesep{0pt}
    \setcellgapes{3pt}\makegapedcells
    \setlength{\tabcolsep}{6.5pt} % Default value: 6pt
    
    \resizebox{0.95\textwidth}{!}{
    \begin{tabular}{r|l|ccc} 
    \hline
    \multirow{2}{*}{
        \begin{tabular}[c]{@{}c@{}}
            Noise\\ 
            ratio
        \end{tabular}
    } & \multirow{2}{*}{Method} & \multicolumn{3}{c}{Accuracy (\%)} \\
	\cmidrule{3-5}
	& & AffectNet & RAF-DB & SFEW \\
	\midrule
	\midrule
    \multirow{4}{*}{10\%} 
     & Baseline & 60.14 $\pm$ 0.23 & 83.28 $\pm$ 0.45  & 45.98 $\pm$ 0.93 \\
     & SCN \cite{scn} & 61.57 $\pm$ 0.15 & 84.65 $\pm$ 0.32  & 49.51 $\pm$ 0.76  \\
     & RUL \cite{rul} & 62.89 $\pm$ 0.13 & 86.24 $\pm$ 0.22  & 47.82 $\pm$ 1.32\\
     & LDLVA (ours) & \textbf{64.37 $\pm$ 0.11} & \textbf{87.98 $\pm$ 0.10}  & \textbf{53.33 $\pm$ 0.57}  \\ 
    \midrule
    \multirow{4}{*}{20\%} 
     & Baseline & 58.37 $\pm$ 0.35 & 81.89 $\pm$ 0.61  & 41.25 $\pm$ 1.12 \\
     & SCN \cite{scn} & 60.83 $\pm$ 0.19 & 83.21 $\pm$ 0.49  & 46.26 $\pm$ 1.24 \\
     & RUL \cite{rul} & 61.74 $\pm$ 0.18 & 84.49 $\pm$ 0.24  & 44.78 $\pm$ 1.04\\ 
     & LDLVA (ours) & \textbf{63.89 $\pm$ 0.14} & \textbf{86.81 $\pm$ 0.12}  & \textbf{51.53 $\pm$ 0.92}  \\  
    \midrule
    \multirow{4}{*}{30\%} 
     & Baseline & 56.94 $\pm$ 0.43 & 78.92 $\pm$ 0.59  & 38.51 $\pm$ 1.69  \\
     & SCN \cite{scn} & 58.80 $\pm$ 0.32 & 80.61 $\pm$ 0.54  & 43.28 $\pm$ 2.06  \\
     & RUL \cite{rul} & 60.77 $\pm$ 0.15 & 82.59 $\pm$ 0.42  & 41.79 $\pm$ 0.81\\
     & LDLVA (ours) & \textbf{62.57 $\pm$ 0.15} & \textbf{85.85 $\pm$ 0.09}  & \textbf{50.3 $\pm$ 0.88}  \\
    \bottomrule
    \end{tabular}
     }
    \caption{Accuracy with synthetic noise. \strut}
    \label{table:noise_result}
    \vspace{-3ex}
\end{table}
The two main aspects of annotation ambiguity in FER are noisy labels and uncertain visual features \cite{scn}. In particular, it can be difficult for people to accurately recognize the emotions on ambiguous facial images, which can result in noisy and incorrect labels. Therefore, we conduct experiments to study the robustness of our LDLVA on mislabelled data by adding synthetic noise to AffectNet, RAF-DB, and SFEW datasets.
% Due to the inherent ambiguity and uncertainty, people may find it difficult to recognize the facial expression correctly, this leads to the noisy and inconsistent annotations in FER datasets. Therefore, the ability to cope with incorrect labels is critical for a successful FER application.
% To that end, we study the robustness of our approach by conducting experiments on RAF-DB, AffectNet and SFEW datasets, in which label noise is synthetically added. 
More specifically, we randomly flip the manual labels to one of the other categories. Three levels of noise are studied in our experiment. We quantitatively evaluate our method and compare with the baseline ResNet-50 \cite{resnet} and recent noise-tolerant FER methods including SCN \cite{scn} and RUL \cite{rul}. 

We perform each experiment three times and report the mean accuracy and standard error in Table \ref{table:noise_result}. The results clearly show that our method consistently outperforms other approaches in all cases. Particularly, our model makes significant improvements over the baseline with an average accuracy margin of 5.13\%, 5.52\%, and 9.81\% on the AffectNet, RAF-DB, and SFEW datasets, respectively. We also observe that the improvements are even more apparent when the noise ratio increases, for example, the accuracy improvement on RAF-DB is 4.7\% with 10\% noise and 6.93\% with 30\% noise. 
% This means our method can effectively disambiguate the noisy label and improve the robustness.
% Moreover, compared to the second best method RUL \cite{rul}, our approach also gains an average improvement of +2.44\%, +1.81\% accuracy across the RAF-DB and AffectNet dataset. Another finding is that RUL \cite{rul} perform worse than SCN \cite{scn} on the SFEW dataset. The reason might be that in RUL training strategy, the network is enforced to recognize the right classes from the mixed representation of a facial image pair. This may not be well suited for small dataset like SFEW since the data is insufficient to learn the pairwise relationship of two different categories. 
The consistent results under various settings demonstrate the ability of our method to effectively deal with noisy annotation, which is crucial in the robustness against label ambiguity. 
\subsection{Experiments with Inconsistent Labels}
\vspace{-3ex}
\begin{table}[h]
\centering
    % make the vertical line continuos
    \setlength\aboverulesep{0pt}\setlength\belowrulesep{0pt}
    \setcellgapes{3pt}\makegapedcells
    \setlength{\tabcolsep}{6.5pt}  % Default value: 6pt

     \resizebox{\textwidth}{!}{
     \begin{tabular}{l|ccc|c} 
    \toprule
    \multirow{2}{*}{Method} & \multicolumn{4}{c}{Accuracy (\%)} \\
	\cmidrule{2-5}
     & AffectNet & RAF-DB & SFEW & Average \\ 
    \midrule
    \midrule
    % Baseline \cite{ldl_alsg} & 57.97 & 81.81 & 52.19 & 63.99 \\
    AIR \cite{air} & 54.23 & 67.37 & 49.88 & 57.16 \\
    NAL \cite{nal} & 55.97 & 84.22 & {58.13} & 66.11 \\
    IPA2LT \cite{ipa2lt} & 57.85 & 83.80 & 53.15 & 64.93 \\
    LDL-ALSG \cite{ldl_alsg} & 58.29 & 85.33 & 55.87 & 66.50 \\ 
    \midrule
    {LDLVA} (ours) & \textbf{62.89} & \textbf{87.26} & \textbf{58.70} & \textbf{69.62}\\
    \bottomrule
    \end{tabular}%
    }
    \caption{Accuracy with inconsistent labels. %Results of previous methods are taken from \cite{ldl_alsg}.
    \strut}
    \label{tab:cross_dataset_result}
    % \vspace{-1ex}
\end{table}

% \begin{table}
% 	\centering
%     \setlength\aboverulesep{0pt}\setlength\belowrulesep{0pt}
%     \setcellgapes{3pt}\makegapedcells
    
%     \setlength{\tabcolsep}{6.5pt}
%     \resizebox{\textwidth}{!}{
%     \begin{tabular}{cc|cc} 
%     % \begin{tabular}{ccc|cc} 
%     \toprule
%     \begin{tabular}[c]{@{}c@{}}Adaptive\\similarity~\end{tabular} & \begin{tabular}[c]{@{}c@{}}Valence-arousal\\value~\end{tabular} &
%     % w/W & w/gVA &
%     \begin{tabular}[c]{@{}c@{}}AffectNet\\(original)~\end{tabular} & \begin{tabular}[c]{@{}c@{}}AffectNet\\(30\% noise)\end{tabular} \\ 
%     \midrule
%     \midrule
%     - & pseudo & 63.95 & 60.52 \\ 
%     \cmark & pseudo & 65.45 & 62.23 \\ 
%     \cmark & groundtruth & 65.74 & 62.57 \\    
%     % \xmark & \xmark & 63.95 & 60.52 \\ 
%     % \cmark & \xmark & 65.45 & 62.23 \\ 
%     % \cmark & \cmark & 65.79 & 62.47 \\
%     \bottomrule
%     \end{tabular}
%     }
%     \caption{Performance of our LDLVA on AffectNet dataset with pseudo vs. groundtruth valence-arousal} 
%     \label{table:ablation_pseudova}
% \vspace{-3ex}
% \end{table}

Due to the ambiguous nature of facial expressions, different individuals can assign different labels for the same image as illustrated in Figure \ref{fig:motivation}. Since the annotations for large-scale FER data are commonly obtained via crowd-sourcing, this can create label inconsistency, especially between different datasets. Therefore, to examine the effectiveness of the proposed methods in dealing with this problem, we follow the cross-dataset protocol in previous state-of-the-art methods \cite{ldl_alsg,ipa2lt} and adopt the experimental settings as proposed in \cite{ldl_alsg} for a fair comparison. 
Specifically, the model is trained using the joint training dataset from  RAF-DB and AffectNet. The resulting model is then tested on all three RAF-DB, AffectNet, and SFEW datasets.

% \begin{table}
% \centering
%     % make the vertical line continuos
%     \setlength\aboverulesep{0pt}\setlength\belowrulesep{0pt}
%     \setcellgapes{3pt}\makegapedcells
%     \setlength{\tabcolsep}{6.5pt}  % Default value: 6pt
    
%     \caption{Test accuracy of different methods with inconsistent labels. %Results of previous methods are taken from \cite{ldl_alsg}.
%     \strut}
%     \begin{tabular}{l|ccc|c} 
%     \toprule
%     \multirow{2}{*}{Method} & \multicolumn{4}{c}{Accuracy (\%)} \\
% 	\cmidrule{2-5}
%      & AffectNet & RAF-DB & SFEW & Average \\ 
%     \midrule
%     \midrule
%     % Baseline \cite{ldl_alsg} & 57.97 & 81.81 & 52.19 & 63.99 \\
%     AIR \cite{air} & 54.23 & 67.37 & 49.88 & 57.16 \\
%     NAL \cite{nal} & 55.97 & 84.22 & {58.13} & 66.11 \\
%     IPA2LT \cite{ipa2lt} & 57.85 & 83.80 & 53.15 & 64.93 \\
%     LDL-ALSG \cite{ldl_alsg} & 58.29 & 85.33 & 55.87 & 66.50 \\ 
%     \midrule
%     % {LDLVA} (ours) & \textbf{62.89} & \textbf{87.26} & 56.61 & \textbf{68.92} \\ 
%     {LDLVA} (ours) & \textbf{60.12} & \textbf{87.26} & \textbf{58.7} & \textbf{68.69}\\
%     \bottomrule
%     \multicolumn{1}{l}{} & \multicolumn{1}{l}{} & \multicolumn{1}{l}{} & \multicolumn{1}{l}{} & \multicolumn{1}{l}{}
%     \end{tabular}%
%     \label{tab:cross_dataset_result}
% \end{table}

Table \ref{tab:cross_dataset_result} reports the results of our experiments. 
Our method achieves the best performance on all three datasets and the highest average accuracy. Notably, LDLVA surpasses the current state-of-the-art LDL-ALSG \cite{ldl_alsg} with an improvement of 3.12\% on average accuracy. Compared to our approach, LDL-ALSG only uses the neighbors' distributions to constrain the network prediction without constructing a label distribution for the center instance. It also lacks a mechanism to adaptively measure the contribution of each neighbor and the uncertainty of the provided annotation. 
The favorable performance confirms the advantages of our method over previous works and demonstrates the generalization ability to data with label inconsistency, which is essential for real-world FER applications.

\subsection{Experiments on Original Datasets}
\label{sec:exp_original}

% ==== merge 3 tables
\begin{table}
\centering
    % make the vertical line continuos
    \setlength\aboverulesep{0pt}\setlength\belowrulesep{0pt}
    \setcellgapes{3pt}\makegapedcells
    \setlength{\tabcolsep}{6.5pt}  % Default value: 6pt

     \resizebox{0.75\textwidth}{!}{
     \begin{tabular}{l|ccc} 
    \toprule
    \multirow{2}{*}{Method} & \multicolumn{3}{c}{Accuracy (\%)} \\
	\cmidrule{2-4}
     & AffectNet & RAF-DB & SFEW \\ 
    \midrule
    \midrule
    Island Loss \cite{island_loss} & - & - & 52.52 \\
    IPFR \cite{IPFR} & 57.40 & - & 55.10 \\
    % gaCNN \cite{gacnn} & 58.78 & 85.07 & - \\
    % DDA \cite{dda_loss} & 62.34 & - & - \\
    EfficientFace \cite{efficientface} & 63.70 & 88.36 & - \\
    % KTN \cite{ktn} & 63.97 & 88.07 & - \\
    DACL \cite{dacl_net} & 65.20 & 87.78 & - \\
    MViT \cite{mvit} & 64.57 & 88.62 & - \\
    RAN	\cite{ran} & - & 86.90 & 56.4 \\
    SCN \cite{scn} & - & 87.03 & - \\
    DMUE \cite{dmue} & - & 88.76 & 57.12 \\
    RUL	\cite{rul} & - & 88.98 & - \\   
    PSR \cite{psr} & 63.37 & 88.98 & - \\
    \midrule
    % {LDLVA} (ours) & \textbf{65.74} \\
    {LDLVA} (ours) & \textbf{66.23} & \textbf{90.51} & \textbf{59.90} \\   
    \bottomrule
    \end{tabular}%
    }
    \caption{Accuracy of different methods on original datasets. \strut} 
    \label{table:clean_data_result}
    \vspace{-3ex}
\end{table}

We further perform experiments on the original AffectNet, RAF-DB, and SFEW to evaluate the robustness of our method to the uncertainty and ambiguity that unavoidably exists in real-world FER datasets. We compare the proposed LDLVA with several state-of-the-art methods in Table \ref{table:clean_data_result}. 
By leveraging label distribution learning on valence-arousal space, our model outperforms other methods and achieves state-of-the-art performance on AffectNet, RAF-DB, and SFEW. Although these datasets are considered to be ``clean", the results suggest that they indeed suffer from uncertainty and ambiguity.
% In addition, to demonstrate that our method can be easily integrated into existing networks, we provide the results corresponding to different backbone architectures in the supplementary. 

\subsection{Qualitative Analysis}
\begin{figure*}[ht]
    \centering
    \vspace{-1ex}
    \includegraphics[width=0.9\textwidth]{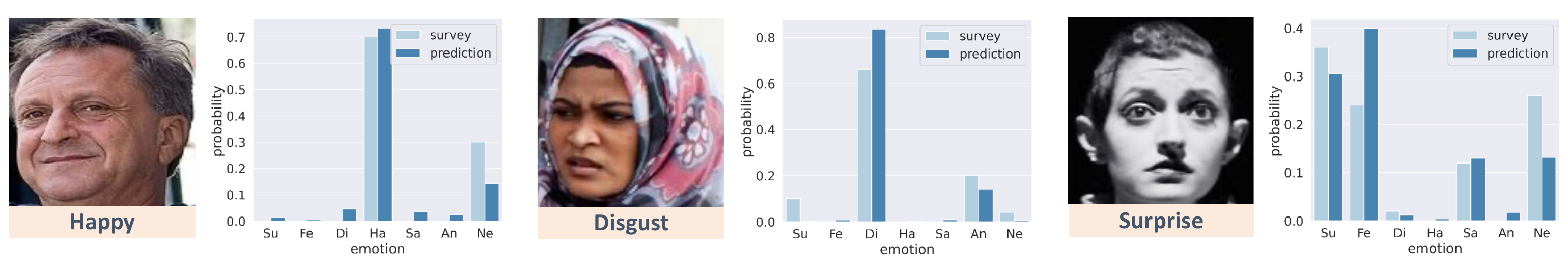}
    \caption{
    Comparison of the results from our survey and our model. More results can be found in the supplementary materials.}
    \vspace{-2ex}
    \label{fig:model_prediction_vs_survey}
\end{figure*}

\textbf{Real-world Ambiguity.} To understand more about real-world ambiguous expressions, we conducted a user study in which we asked 50 participants to choose the most clearly expressed emotion on random test images from RAF-DB and AffectNet datasets. The numbers of votes per class are normalized to obtain the emotion distribution. We compare our model's predictions with the survey results in Figure \ref{fig:model_prediction_vs_survey}.
% The normalized result indicates the probability or the importance degree of each emotion presented in each image. 
% In Figure \ref{fig:model_prediction_vs_survey}, We compare the predicted outputs of our LDLVA model and the survey results. \textit{More qualitative results can be found in the supplementary material}.
We can see that these images are ambiguous as they express a combination of different emotions, hence the participants do not fully agree and have different opinions about the most prominent emotion on the faces. 
% Qualitatively, we observe that ambiguities exist in those images. These ambiguities may exhibit difficulties for people in determining the correct emotion, as we can see the participants do not fully agree with each other. 
It is further shown that LDLVA can give consistent results and agree with the perception of humans to some degree, which suggests that our model can effectively address the ambiguity problem in facial expressions.

\begin{figure*}[!ht]
    \centering
    %\vspace{-3ex}
    \includegraphics[width=0.85\textwidth]{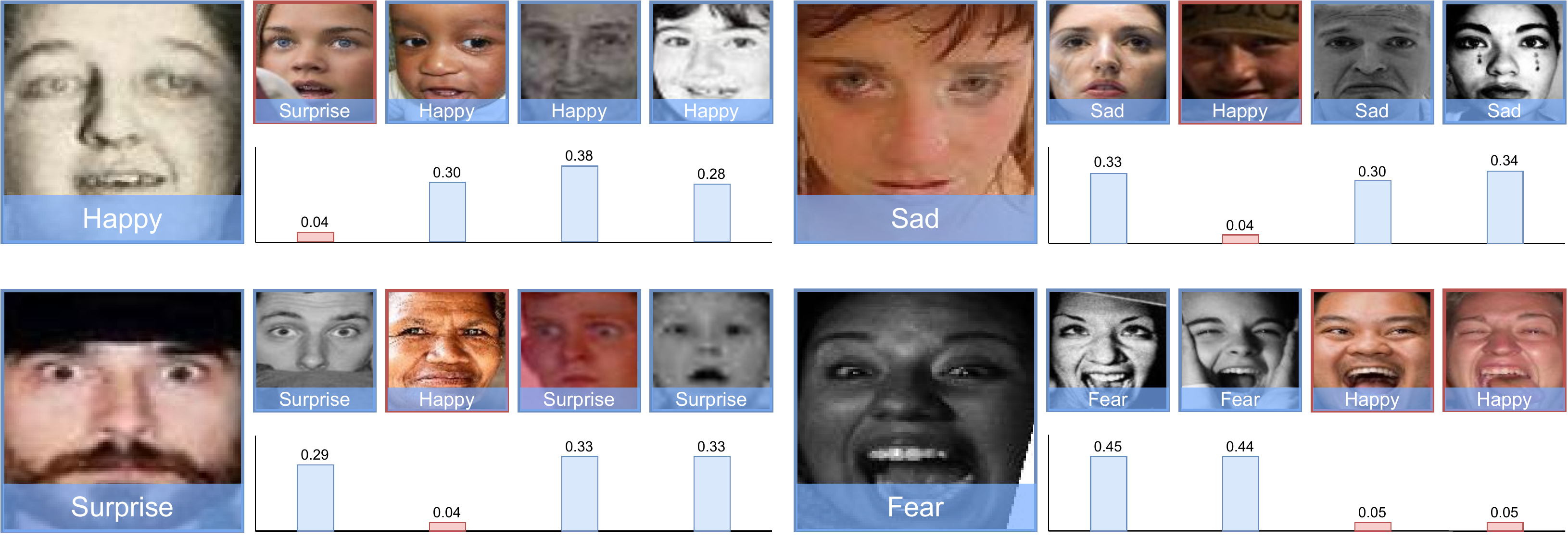}
    \caption{The calibration scores of neighbor images with respect to the main instance. The large image on the left is the main instance. The neighbor images are shown at the top, and their corresponding scores are shown at the bottom.}
    \label{fig:visualize_weighting}
    % \vspace{-1ex}
\end{figure*}

\textbf{Adaptive Similarity.} Figure \ref{fig:visualize_weighting} presents the normalized {calibration scores} of different neighbors with respect to the center instance computed by our adaptive similarity mechanism. It can be seen that some neighbors may look visually similar to the central image but do not express the same emotion. By giving low calibration values, our method can effectively suppress the negative influence of these neighbors and lower their contribution, hence resulting in a more robust and accurate estimation of the emotion distribution. 
% On the other hand, the Weighting Module does not completely omit the contribution of these neighbors. % Since they share similar characteristics with the central image, the model can utilize that information to learn more robust features. 

\begin{figure*}[!ht]
    \centering
    \includegraphics[width=0.85\textwidth]{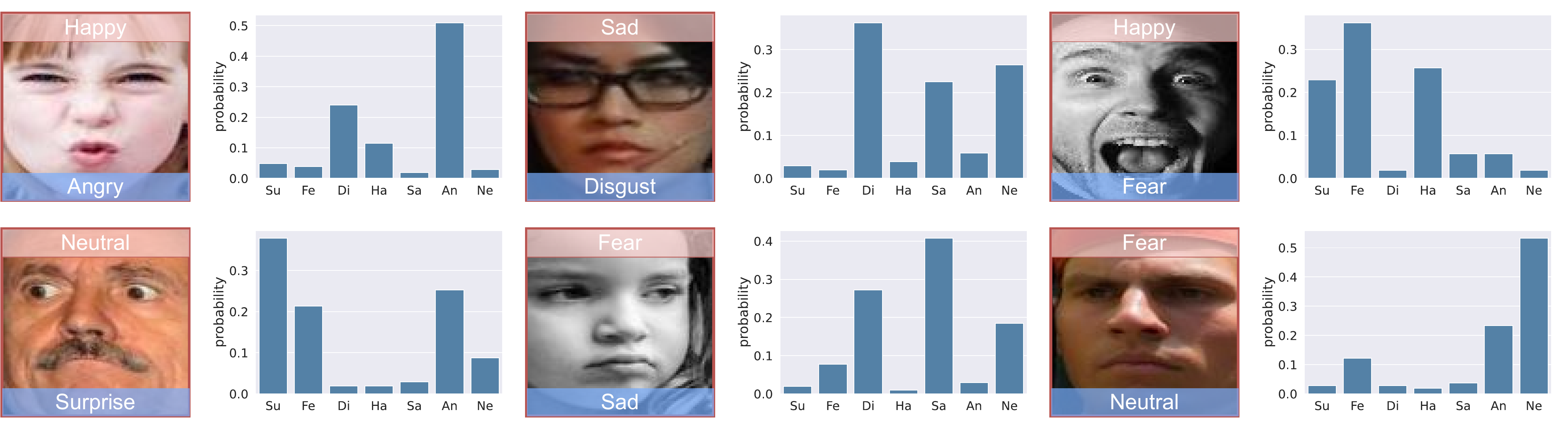}
    \caption{Examples of the emotion distribution recovered by our method when the dataset is contaminated with noisy labels. The label on top of each image is the synthetic noisy label and the bottom denotes the human annotation.}
    \label{fig:visualize_dist}
\end{figure*}

\begin{figure}[h]
    \centering
    \vspace{-2ex}
    \includegraphics[width=0.96\textwidth]{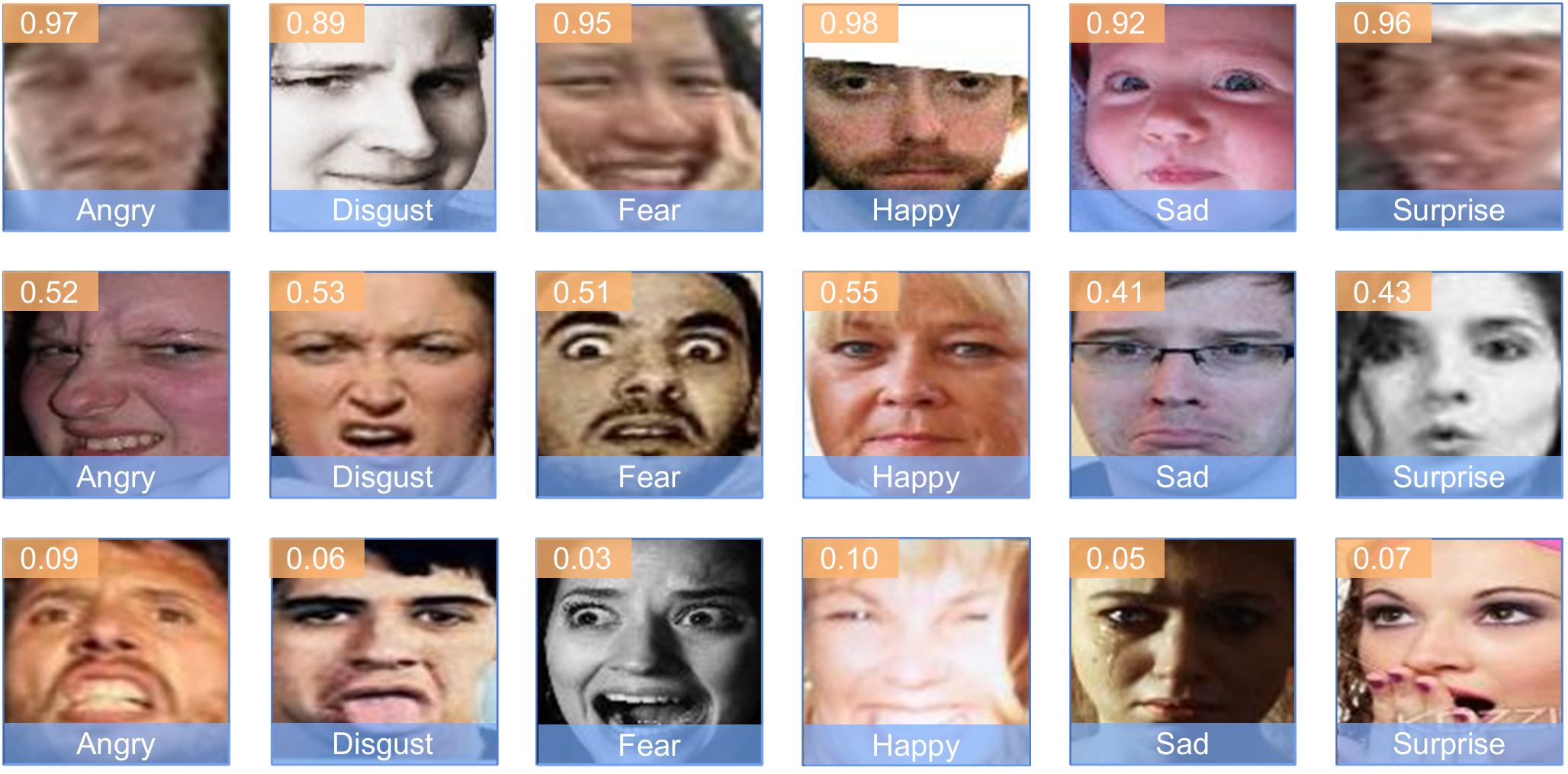}
    \caption{
    Visualization of uncertainty values of some examples from RAF-DB dataset. %Each image is tagged with its estimated uncertainty values and the original label. High uncertainty value means that the model is uncertain about the label of an image and vice versa.
    }
    \label{fig:uncertainty}
\end{figure}
% \vspace{-3ex}

\textbf{Constructed Label Distribution.} %To further investigate the effectiveness of our framework in addressing the ambiguity problem in emotion recognition, 
In Figure \ref{fig:visualize_dist}, we visualize the emotion distributions reconstructed by our method on mislabelled images. Despite the incorrect annotations, our approach is able to construct plausible distributions and discover the correct labels. It is also noteworthy that some expressions manifest multiple emotions rather than only the single provided category, which means the discovered distribution can provide more supervision for training. 
% Furthermore, we can observe that ambiguity exists in these images. For instance, the facial images are mixed with multiple expressions and a single discrete label is insufficient to represent these emotions. %As a consequence, these weaknesses may hinder the FER performance and other related emotion analysis tasks. 
% We can see that it is not a trivial task to quantify the emotion of these images correctly. The qualitative results show that our method is not only robust to the noisy label but is also able to handle the ambiguity problem in FER.

\textbf{Uncertainty Factor.} 
% We further qualitatively investigate our method under the uncertainties that inevitably exists in real-world FER datasets. 
Figure \ref{fig:uncertainty} shows the estimated uncertainty factors of some training images in the RAF-DB dataset and their original labels. The uncertainty values decrease from top to bottom. Highly uncertain labels can be caused by low-quality inputs (as shown in \texttt{Angry} and \texttt{Surprise} columns) or ambiguous facial expressions. In contrast, when the emotions can be easily recognized as those in the last row, the uncertainty factors are assigned low values. This characteristic can guide the model to decide whether to put more weights on the provided label or the neighborhood information. Therefore, the model can be more robust against uncertainty and ambiguity.
% In the first row, we visualize images with large uncertainty values. We can see that these uncertainties are attributed to various factors, such as the low-quality images or ambiguous expressions. 
% These factors can lead to inconsistent and noisy annotations, which may potentiality hinder the ability of a FER system if they are not considered properly. In contrast, the facial expressions in the last row are apparently recognizable, which leads to low uncertainty values as the model are more confident about the expression label of these images. These analyses demonstrate the benefits of our uncertainty-aware scheme as it help mitigate the adverse effect of ambiguous data. This can aid the model to appropriately reconstruct the emotion distribution since it is encouraged to pay more attention to the neighbor information.

\subsection{Ablation Study}
% combine ablation analysis and weighting module table into a single analysis
\textbf{Contribution of 
Each Component.} In Table \ref{tab:ablation_study}, we present the accuracy corresponding to different combinations of our components: label distribution (whether to construct $\bm{d}^i$ or not), adaptive similarity (whether to compute calibration scores or directly use local similarity scores as contribution degrees), uncertainty factor (whether to use separate $\lambda^i$ for each instance or share a fixed value $\lambda$ for all training samples), and discriminative loss (whether to incorporate $\mathcal{L}_D$ in Equation \ref{eq:total_loss} or not). By employing label distribution with adaptive similarity \texttt{(ii)}, we can significantly improve the accuracy of the vanilla approach \texttt{(i)} by 1.89\% on original RAF-DB and 3.77\% on 30\%-noise RAF-DB. Further integrating uncertainty factor and discriminative loss consistently boost the performance of the model, as shown in the results of \texttt{(iii)} and \texttt{(v)}, respectively. 
%We also note that without adaptive similarity as in \texttt{(iv)}, the accuracy of the model drops from 89.57\% and 85.85\% to 88.52\% and 83.56\% on the original and 30\%-noise RAF-DB. 
The results show the effectiveness of each component as well as the advantages of their combination in our LDLVA method. 

% \begin{table}
% 	\centering
%     \setlength\aboverulesep{0pt}\setlength\belowrulesep{0pt}
%     \setcellgapes{3pt}\makegapedcells
    
%     \setlength{\tabcolsep}{6.5pt} % Default value: 6pt
%     \renewcommand{\arraystretch}{1.5} % Default value: 1
%     \caption{Ablation analysis of different components in our method. \strut} 
%     \resizebox{\textwidth}{!}{
%     \begin{tabular}{c|cccc|cc} 
%     \midrule
%     Setting & 
%     \begin{tabular}[c]{@{}c@{}}Label\\distribution~\end{tabular} &
%     \begin{tabular}[c]{@{}c@{}}Adaptive\\similarity~\end{tabular} &    \begin{tabular}[c]{@{}c@{}}Uncertainty\\factor~\end{tabular} &
%     \begin{tabular}[c]{@{}c@{}}Discriminative\\loss~\end{tabular} & 
%     \begin{tabular}[c]{@{}c@{}}RAF-DB\\(original)~\end{tabular} & \begin{tabular}[c]{@{}c@{}}RAF-DB\\(30\% noise)\end{tabular} \\ 
%     \midrule
%     \midrule
%     \texttt{(i)} & - & - & - & - & 86.27 & 78.92 \\ 
%     \midrule
%     \texttt{(ii)} & \cmark & \cmark & - & - & 88.16 & 82.69 \\ 
%     \texttt{(iii)} & \cmark & \cmark & \cmark & - & 89.28 & 84.38 \\    
%     \midrule
%     \texttt{(iv)} & \cmark & - & \cmark & \cmark & 88.52 & 83.56 \\   
%     \texttt{(v)} & \cmark & \cmark & \cmark & \cmark & 89.57 & 85.85 \\     
%     \bottomrule
%     \end{tabular}
%     }
%     \label{table:ablation_study}
%     \vspace{-3ex}
% \end{table}

\begin{table}
\centering
    % make the vertical line continuos
    \setlength\aboverulesep{0pt}\setlength\belowrulesep{0pt}
    \setcellgapes{3pt}\makegapedcells
    \setlength{\tabcolsep}{6.5pt}  % Default value: 6pt
    \resizebox{\textwidth}{!}{%
    \begin{tabular}{c|cccc|cc} 
    \midrule
    Setting & 
    \begin{tabular}[c]{@{}c@{}}LD~\end{tabular} &
    \begin{tabular}[c]{@{}c@{}}AS~\end{tabular} &    \begin{tabular}[c]{@{}c@{}}UF~\end{tabular} &
    \begin{tabular}[c]{@{}c@{}}DL~\end{tabular} & 
    \begin{tabular}[c]{@{}c@{}}RAF-DB\\(original)~\end{tabular} & \begin{tabular}[c]{@{}c@{}}RAF-DB\\(30\% noise)\end{tabular} \\ 
    \midrule
    \midrule
    \texttt{(i)} & - & - & - & - & 87.06 & 78.92 \\ 
    \midrule
    \texttt{(ii)} & \cmark & \cmark & - & - & 88.95 & 82.69 \\ 
    \texttt{(iii)} & \cmark & \cmark & \cmark & - & 89.57 & 84.38 \\    
    \midrule
    \texttt{(iv)} & \cmark & - & \cmark & \cmark & 89.31 & 83.56 \\ 
    \texttt{(v)} & \cmark & \cmark & \cmark & \cmark & 90.51 & 85.85 \\     
    \bottomrule
    \end{tabular}
    }
\caption{Component analysis (LD: Label Distribution, AS: Adaptive Similarity, UF: Uncertainty Factor, DL: Discriminative Loss) \strut}
\label{tab:ablation_study}
\vspace{-3ex}
\end{table}

\begin{figure}
\vspace{-3ex}
  {
  \centering
  \includegraphics[width=0.75\textwidth]{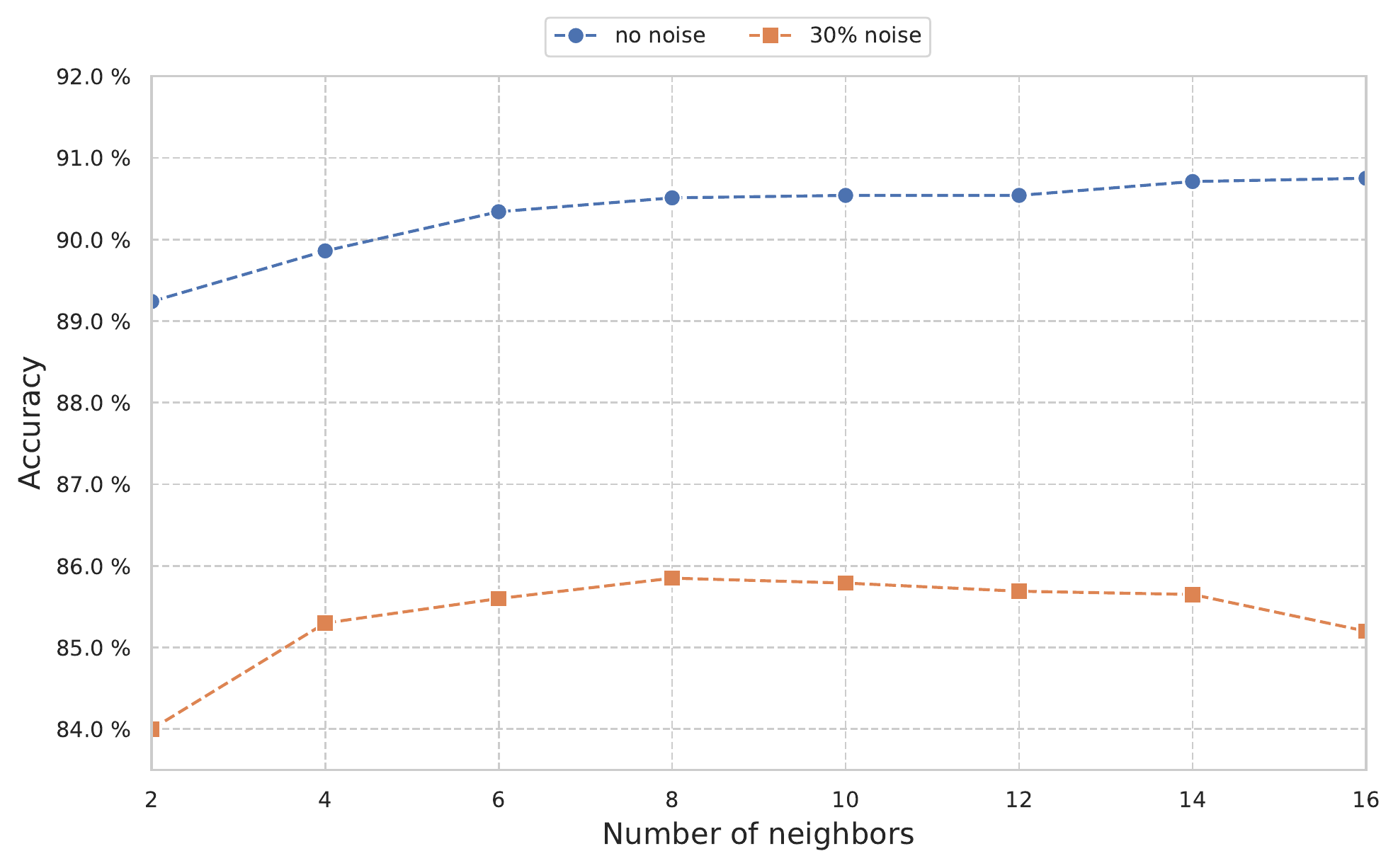}
 }
  {\caption{Evaluation results with different numbers of neighbors.}
  \label{fig:ablation_study}}
\vspace{-3ex}
\end{figure}

\textbf{Number of Nearest Neighbors.} We present the effect of the number of nearest neighbors $K$ on the model performance in Figure {\ref{fig:ablation_study}}. For original RAF-DB data, higher values of $K$ give better results but also require more training time. In particular, our training time with $K=8$ and $K=16$ on AffectNet is 12 hours and 20 hours, respectively. Under noisy conditions, the best result is obtained with $K=8$ while larger or smaller $K$ can lead to slightly worse performance. The reason is that using a large $K$ might include more corrupted labels while using too few neighbors can limit the amount of exploitable information.

\section{Conclusion}
\label{sec:Conclusion}
This paper introduces a new label distribution learning method for facial expression recognition by leveraging structure information in the valence-arousal space to recover the intensities distributed over emotion categories. 
% We have achieved that by proposing a framework with the Weighting Module and Distribution Aggregator to reconstruct the emotion distribution of the main instance based on its neighborhood. 
We first employ the adaptive similarity to account for the errors caused by pseudo valence-arousal and robustly measure the contribution degree of each neighbor. Then, the target label distribution is constructed by incorporating both the provided single label and the combination of neighbor distribution guided by the uncertainty value. The constructed label distribution provides rich information about the emotions, thus can effectively describe the ambiguity degree of the facial image. %Furthermore, We also identify the existing limitations and propose to modify the standard center loss so that it can increase the robustness and discriminative of the learned features, especially in ambiguous scenarios.
Intensive experiments on popular datasets demonstrate the effectiveness of our method over previous approaches under inconsistency and uncertainty conditions in facial expression recognition. %Our source code and trained model will be released for further study.

\textbf{Acknowledgement.}
The authors want to express our appreciation to Dr. Thieu Vo for his valuable feedback during the preparation of this paper. We also want to thank all volunteers that participated in our user study.

% \clearpage\mbox{}Page \thepage\ of the manuscript.
% \clearpage\mbox{}Page \thepage\ of the manuscript.

% This is the last page of the manuscript.
% \par\vfill\par
% Now we have reached the maximum size of the ECCV 2022 submission (excluding references).
% References should start immediately after the main text, but can continue on p.15 if needed.

{\small
\bibliographystyle{ieee_fullname}
\bibliography{egbib}
}

\end{document}

% \renewcommand\thelinenumber{\color[rgb]{0.2,0.5,0.8}\normalfont\sffamily\scriptsize\arabic{linenumber}\color[rgb]{0,0,0}}
% \renewcommand\makeLineNumber {\hss\thelinenumber\ \hspace{6mm} \rlap{\hskip\textwidth\ \hspace{6.5mm}\thelinenumber}}
% \linenumbers
\pagestyle{headings}
\mainmatter
\def\ECCVSubNumber{2742}  % Insert your submission number here

\title{Supplementary Materials for \\ Uncertainty-aware Label Distribution Learning for Facial Expression Recognition} 

% INITIAL SUBMISSION 
%\begin{comment}
\titlerunning{ECCV-22 submission ID \ECCVSubNumber} 
\authorrunning{ECCV-22 submission ID \ECCVSubNumber} 
\author{Anonymous ECCV submission}
\institute{Paper ID \ECCVSubNumber}
%\end{comment}
%******************

% CAMERA READY SUBMISSION
\begin{comment}
\titlerunning{Abbreviated paper title}

\author{First Author\inst{1}\orcidID{0000-1111-2222-3333} \and
Second Author\inst{2,3}\orcidID{1111-2222-3333-4444} \and
Third Author\inst{3}\orcidID{2222--3333-4444-5555}}
%
\authorrunning{F. Author et al.}
% First names are abbreviated in the running head.
% If there are more than two authors, 'et al.' is used.
%
\institute{Princeton University, Princeton NJ 08544, USA \and
Springer Heidelberg, Tiergartenstr. 17, 69121 Heidelberg, Germany
\email{lncs@springer.com}\\
\url{http://www.springer.com/gp/computer-science/lncs} \and
ABC Institute, Rupert-Karls-University Heidelberg, Heidelberg, Germany\\
\email{\{abc,lncs\}@uni-heidelberg.de}}
\end{comment}
%******************
\maketitle

\section{Experiments on Original Datasets}
\begin{table}[]
    \centering
    \setlength\aboverulesep{0pt}\setlength\belowrulesep{0pt}
    \setcellgapes{3pt}\makegapedcells
    \setlength{\tabcolsep}{7pt}  
    
    \begin{tabular}{l|c|cc} 
    \toprule
    \multirow{2}{*}{Backbone} &
    \multirow{2}{*}{LDLVA} &
    \multicolumn{2}{c}{Accuracy (\%)} \\
	\cmidrule{3-4}
     & & RAF-DB & AffectNet \\ 
    \midrule
    \midrule
    MobileNetV2 & - &  85.45 & 61.32 \\
    MobileNetV2 & \cmark & 88.62 & 65.06 \\
    \midrule
    ResNet-18 & - & 86.27 & 62.24 \\
    ResNet-18 & \cmark & 89.57 & 65.74 \\
    \midrule
    ResNet-50 & - & 87.06 & 63.19 \\
    ResNet-50 & \cmark & 90.51 & 66.23 \\
    \midrule 
    ResNet-101 & - & 88.28 & 63.50 \\ 
    ResNet-101 & \cmark & 91.26 & 66.48 \\
    \bottomrule
    \end{tabular}
    \caption{Test accuracy of different backbone architectures on original RAF-DB and AffectNet datasets.}
    \label{tab:different_backbones}
\end{table}

% \begin{table}[]
%     \centering
%     \setlength\aboverulesep{0pt}\setlength\belowrulesep{0pt}
%     \setcellgapes{3pt}\makegapedcells
%     \setlength{\tabcolsep}{7pt}  
    
%     \begin{tabular}{l|c|c|cc} 
%     \toprule
%     \multirow{2}{*}{Backbone} &
%     \multirow{2}{*}{LDLVA} &
%     \multirow{2}{*}{Params} &
%     \multicolumn{2}{c}{Accuracy (\%)} \\
% 	\cmidrule{4-5}
%      & & & RAF-DB & AffectNet \\ 
%     \midrule
%     \midrule
%     MobileNetV2 & - &  & 85.45 & 61.32 \\
%     MobileNetV2 & \cmark & & 88.62 & 65.06 \\
%     \midrule
%     ResNet-18 & - & & 86.27 & 62.24 \\
%     ResNet-18 & \cmark & & 89.57 & 65.74 \\
%     \midrule
%     ResNet-50 & - & & 87.06 & 63.19 \\
%     ResNet-50 & \cmark & & 90.51 & 66.23 \\
%     \midrule 
%     ResNet-101 & - & & 88.28 & 63.50 \\ 
%     ResNet-101 & \cmark & & 91.26 & 66.48 \\
%     \bottomrule
%     \end{tabular}
%     \caption{Test accuracy of different backbone architectures on original RAF-DB and AffectNet datasets.}
%     \label{tab:different_backbones}
% \end{table}

To demonstrate that our method can be easily integrated into existing networks to enhance the robustness, we provide the results corresponding to different backbone architectures. Specifically, we apply LDLVA to the MobileNetV2~\cite{mobilenetv2}, ResNet-18, ResNet-50, and ResNet-101~\cite{resnet} and report the results on RAF-DB and AffectNet datasets in Table. \ref{tab:different_backbones}. The results show that LDLVA consistently improves the performance of all architecture by a large margin. In addition, we observe that there is a trade-off between the architecture complexity and the computational costs. More complicated architecture can give better performance but requires more memory and computations. This suggests that choosing a suitable backbone architecture is also important for various FER applications.

\section{Visualization of Learned Features}
\begin{figure*}
\captionsetup[subfigure]{labelformat=empty, farskip=1.5pt}
\centering
\begin{tabular}{cccc}
\rotatebox{90}{Original} &
\subfloat{\includegraphics[width = 0.3\textwidth]{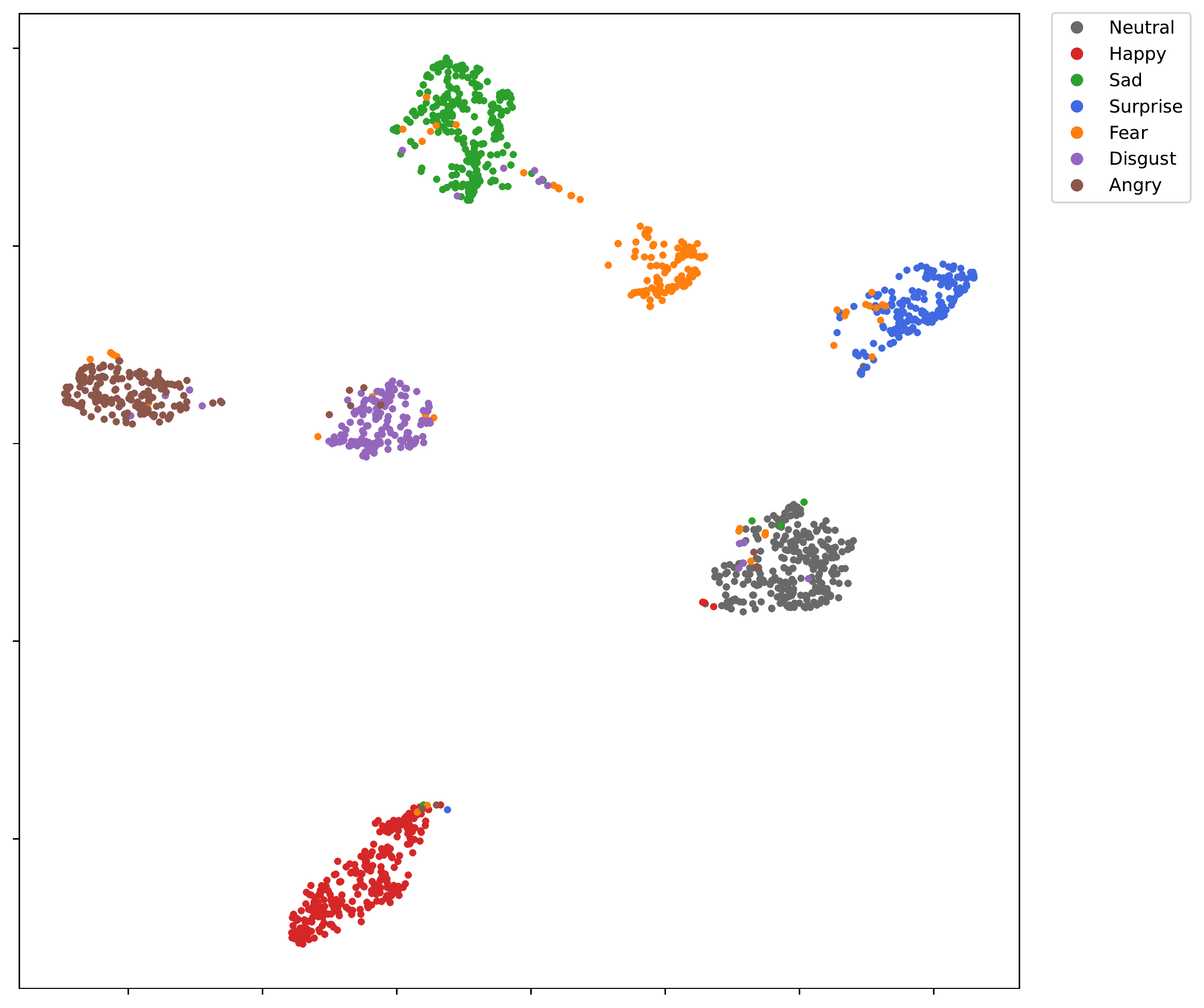}} &
\subfloat{\includegraphics[width = 0.3\textwidth]{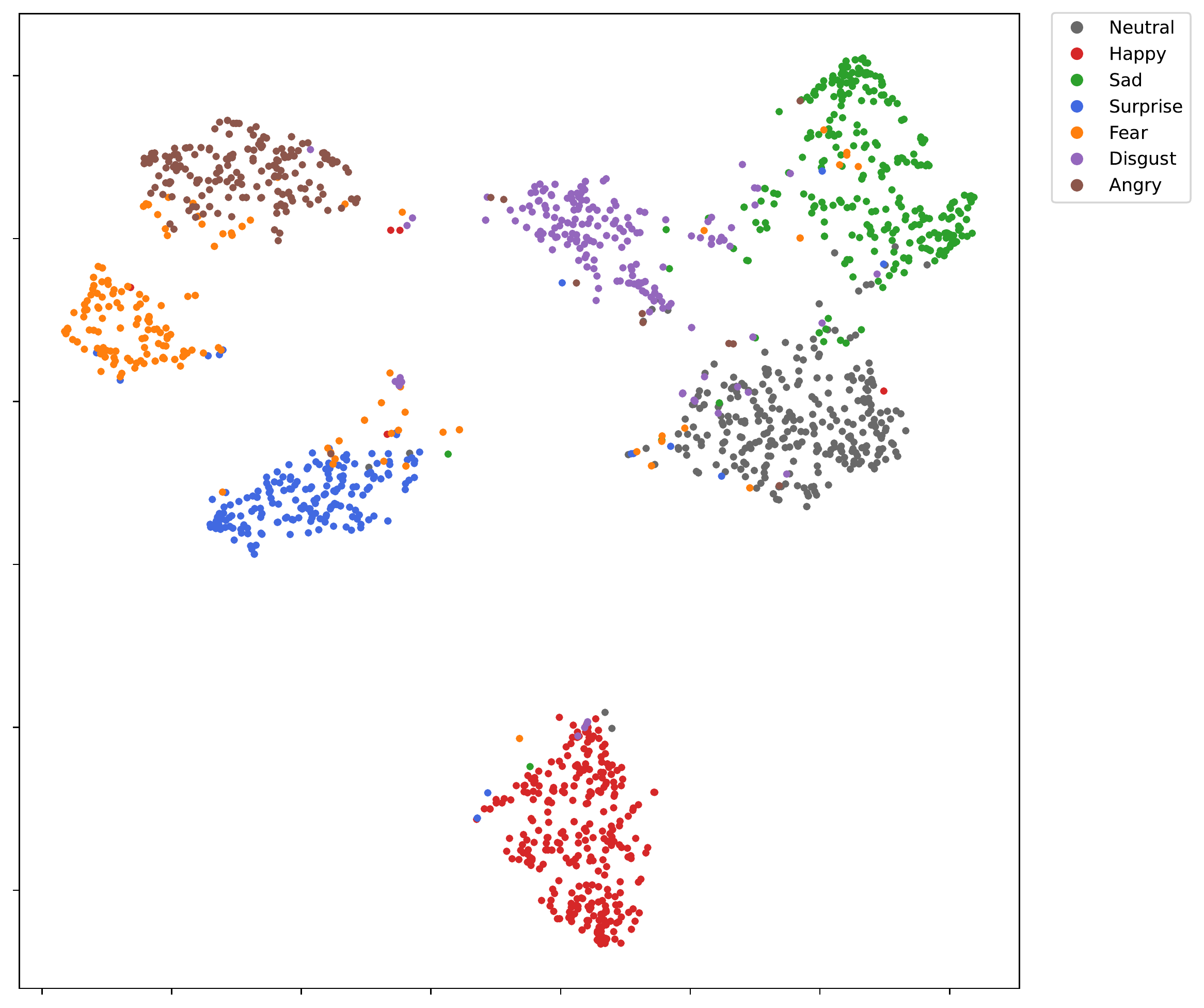}} &
\subfloat{\includegraphics[width = 0.3\textwidth]{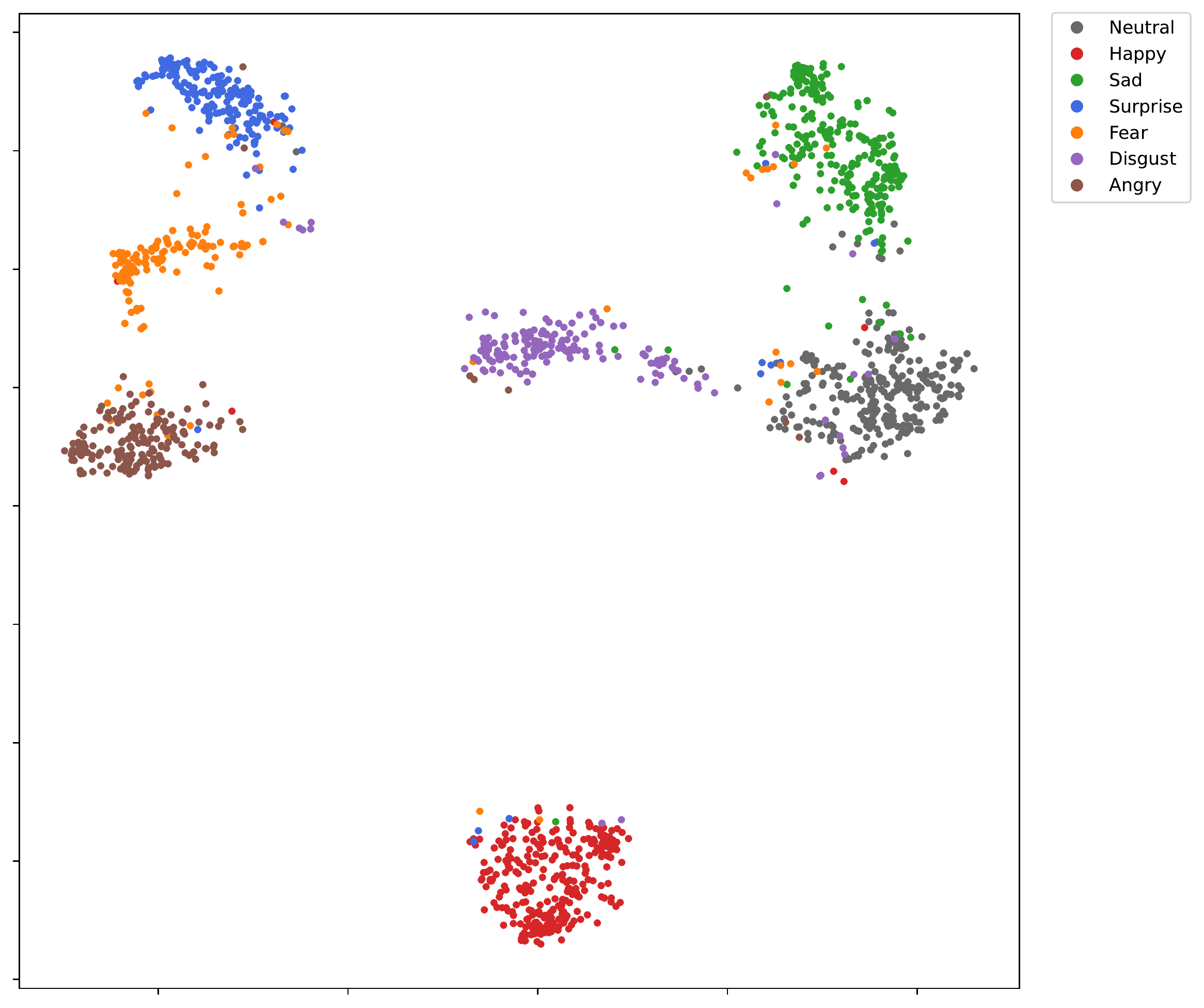}} \\
\rotatebox{90}{30\% noise} & 
\subfloat[(a) Discriminative loss]{\includegraphics[width=0.3\textwidth]{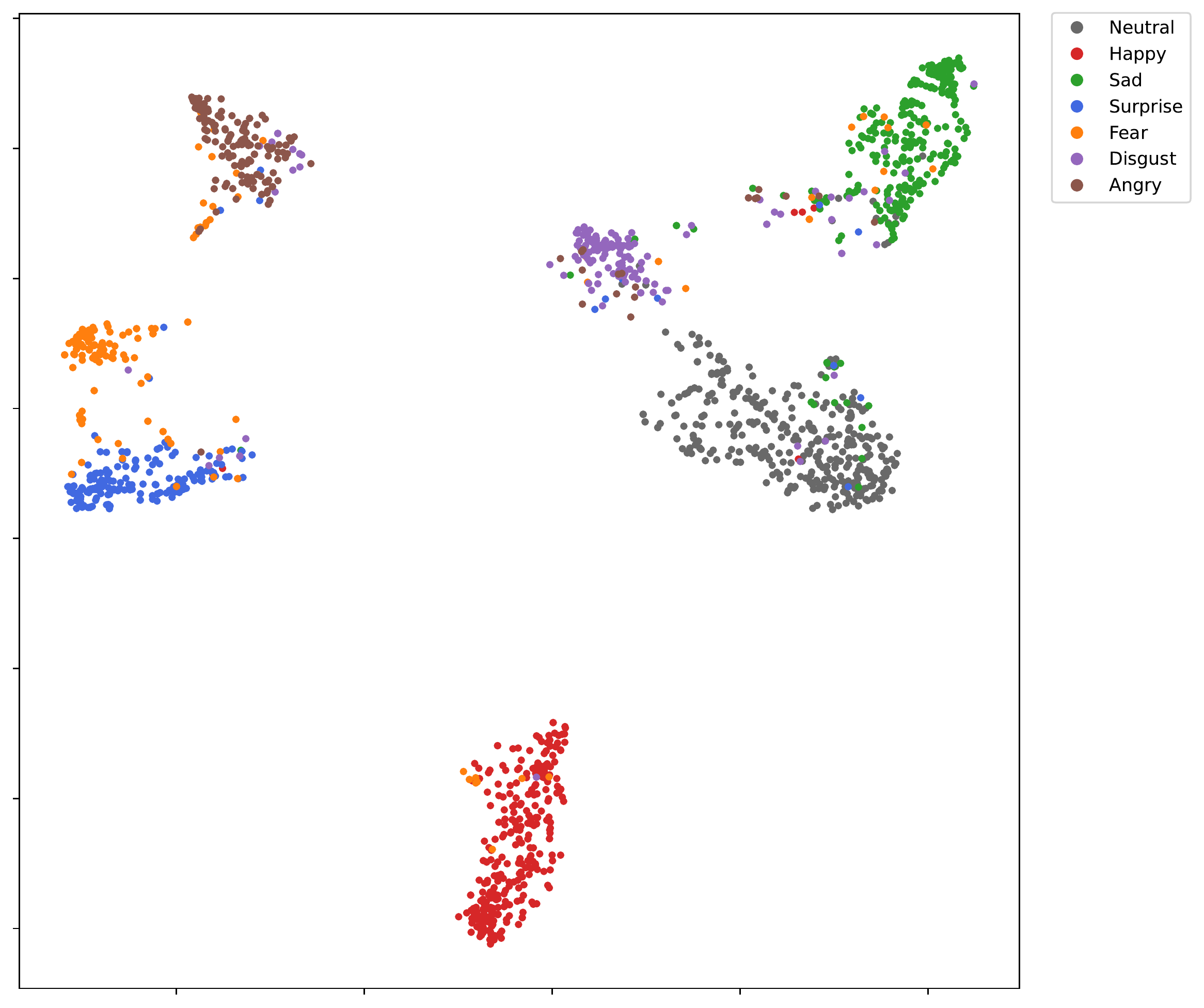}} &
\subfloat[(b) No loss]{\includegraphics[width = 0.3\textwidth]{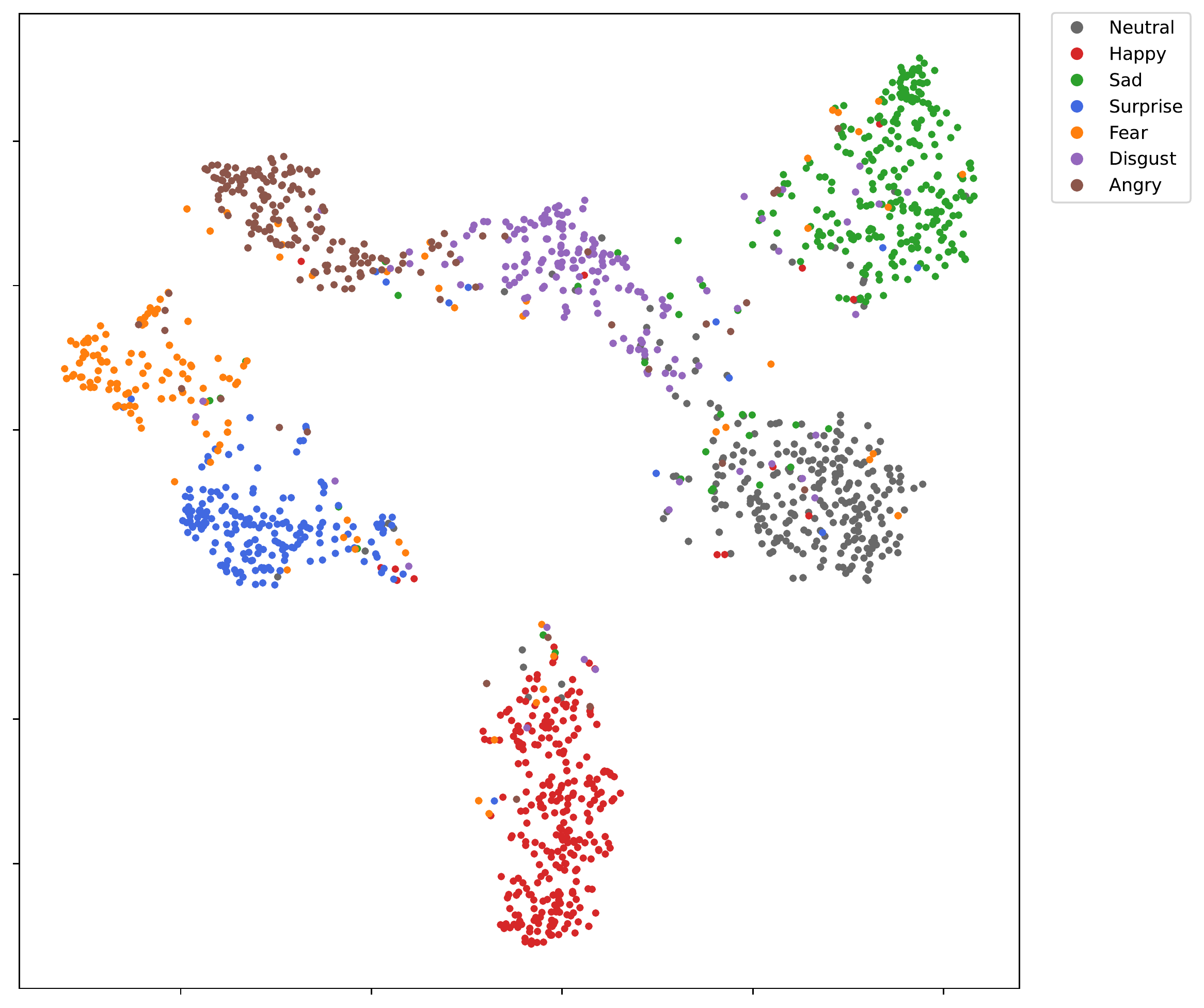}} &
\subfloat[(c) Center loss]{\includegraphics[width = 0.3\textwidth]{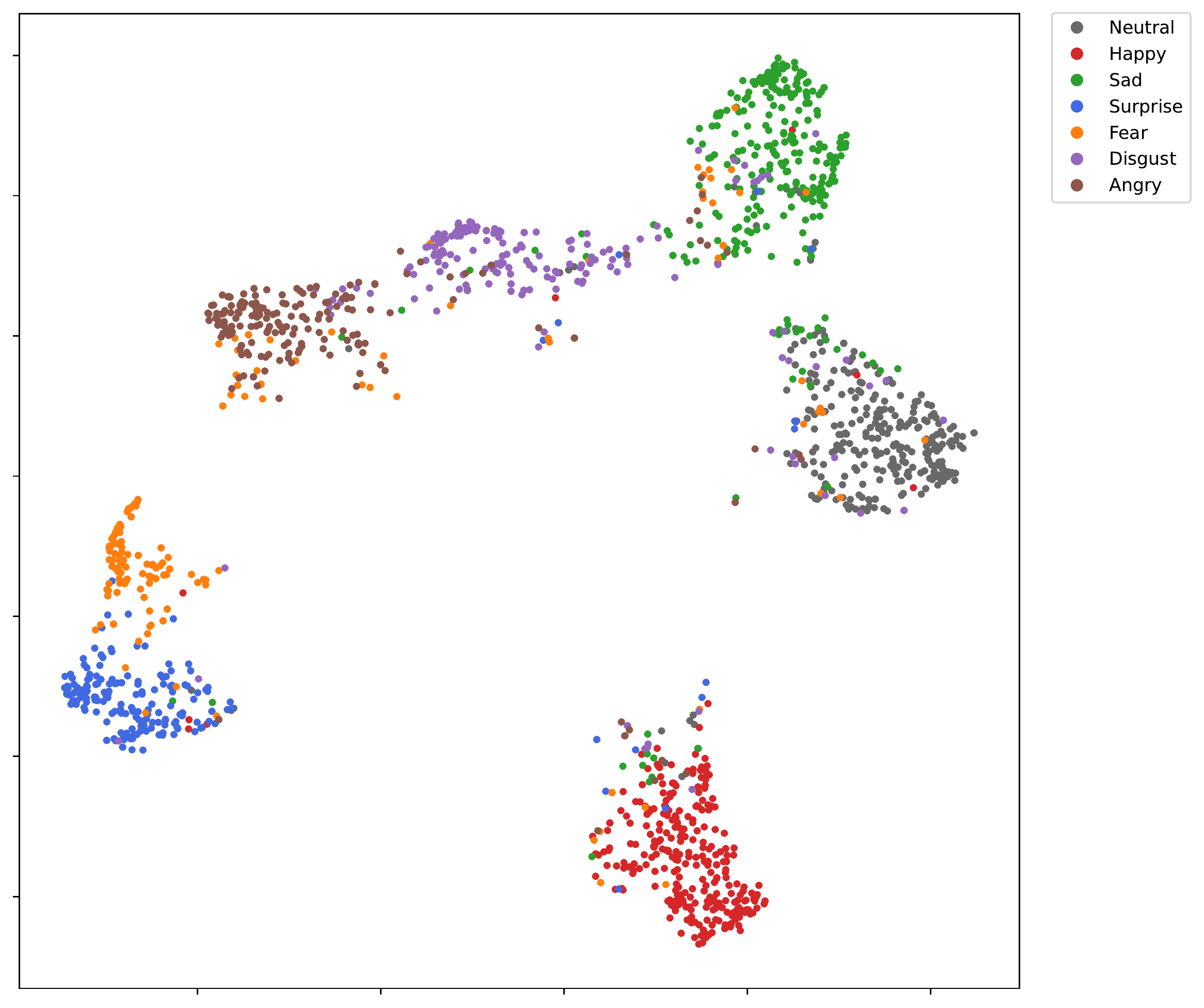}}\\
\end{tabular}
\vspace{0ex}
\caption{Visualization of the learned features using (a) our discriminative loss, (b) no loss, and (c) center loss on random samples from RAF-DB dataset}
\label{fig:vis_features}
\end{figure*}

In Figure \ref{fig:vis_features}, we visualize the t-SNE~\cite{tsne} embedding of the learned features of (a) our discriminative loss, (b) without using additional loss function, and (c) center loss on randomly chosen samples from the RAF-DB dataset.  Although the center loss (c) makes the features belonging to the same class closer to each other, there still exists overlappings between clusters. This problem is even more obvious on the dataset with noisy labels, since the features on mislabelled examples are pulled toward their wrong labels. In contrast, incorporating our proposed discriminative loss results in more compact and better separated clusters. It can also help mitigate the effect of the noisily learned features. 
This indicates that learning to discriminate ambiguous facial features is also important for robust facial expression recognition.

\section{User Survey for Real-world Ambiguity}
As described in the main paper, we randomly picked 21 images from FER test sets and asked 50 volunteers to select the most clearly expressed emotion on each image. We then compare the results of our proposed method and the current state-of-the-art method DMUE \cite{dmue} on our user survey data. Quantitatively, we use Jeffrey's divergence to measure the difference between the emotion distributions voted by human and predicted by each model. The formula is as follows:
\begin{equation}
    D = \frac{1}{n} \sum_{i=1}^n D_J\left(f(\bm{x}^i; \theta), \bm{d}^i\right) = \frac{1}{n} \sum_{i=1}^n \sum_{j=1}^m \left(d^i_{y_j} - f_j(\bm{x}^i; \theta)\right)\log\frac{d^i_{y_j}}{f_j(\bm{x}^i; \theta)}
\end{equation}

As shown in Table \ref{tab:survey_result}, it is clear that the predictions of our method are significantly more similar to human perceptions than those of DMUE \cite{dmue}, demonstrated by a much lower Jeffrey's divergence value. 

\begin{table}[]
    \centering
    \setlength\aboverulesep{0pt}\setlength\belowrulesep{0pt}
    \setcellgapes{3pt}\makegapedcells
    \setlength{\tabcolsep}{7pt}  
    \begin{tabular}{l|c}
    \toprule
    Method & Jeffrey's divergence \\
    \midrule
    \midrule
    DMUE & 4.80 \\
    LDLVA (ours) & 1.74 \\
    \bottomrule
    \end{tabular}
    \caption{Average Jeffrey's divergence of our method and DMUE \cite{dmue} on survey data.}
    \label{tab:survey_result}
\end{table}

\begin{figure*}[ht]
    \centering
    \vspace{-1ex}
    \includegraphics[width=0.99\textwidth]{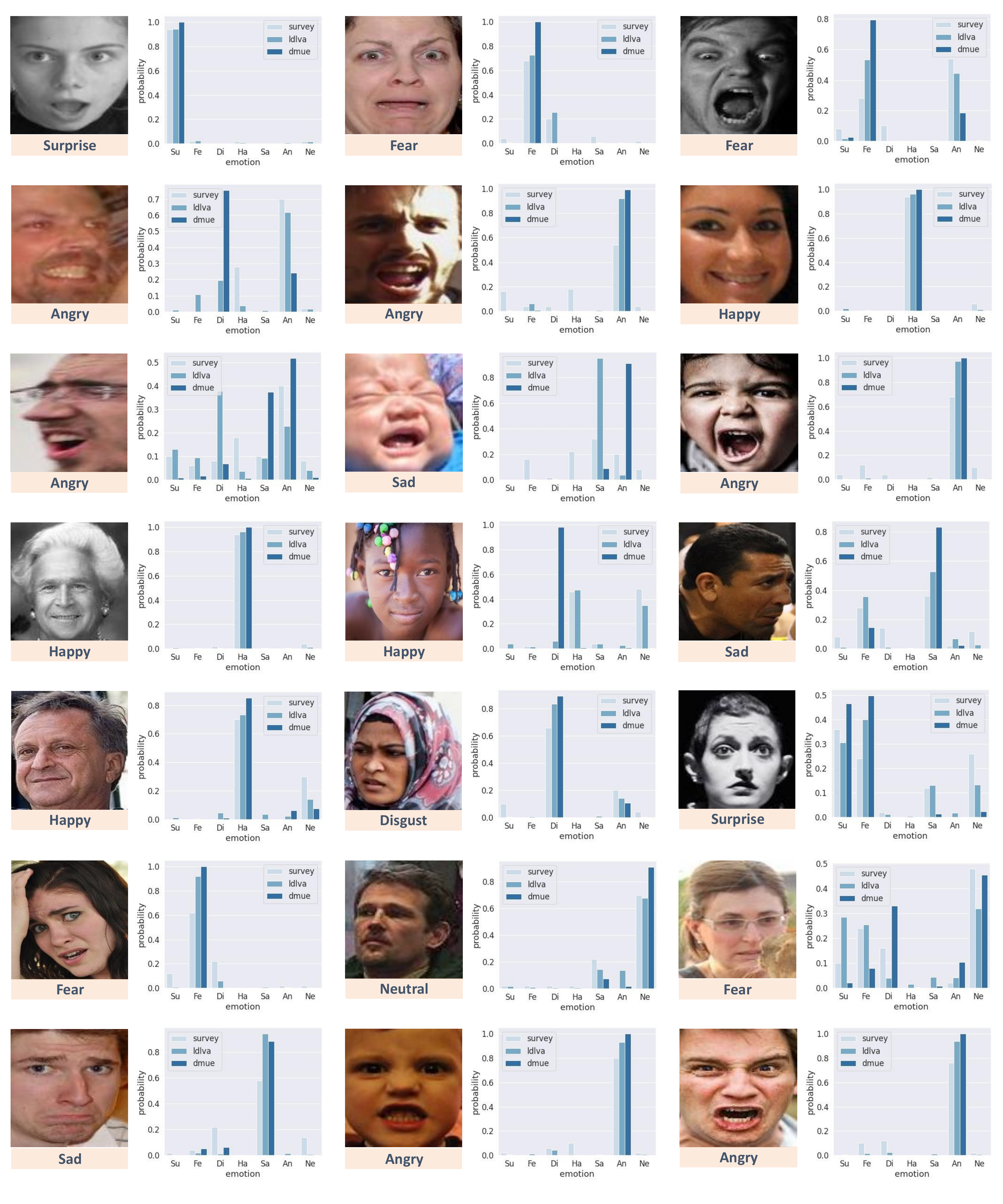}
    \caption{
    Comparison of the emotion distributions of the user survey, our LDLVA model and DMUE \cite{dmue} method.}
    \label{fig:survey_result}
\end{figure*}

Furthermore, we also compare the two models qualitatively and present the results in Figure \ref{fig:survey_result}. Generally, although the most confident emotion of both models matches with the survey, we can see that the distributions predicted by our model follow human results more closely than the previous method DMUE. While our model tends to distribute scores for different emotions, especially in ambiguous cases, DMUE model gives very high scores for only 1 or 2 classes. This can lead to the case when DMUE fails to cover the expressed emotions (the middle image in the fourth row, for example) but our LDLVA can give the correct prediction. 

These results confirm that training models directly with label distributions as in our proposed approach can effectively help to cover multiple emotions that are often expressed in real-world scenarios.   

\clearpage
\bibliographystyle{splncs04}
\bibliography{egbib}